\documentclass[runningheads]{llncs}
 
\usepackage[mobile]{eccv}

\usepackage{eccvabbrv}

\usepackage{graphicx}
\usepackage{amsmath}
\usepackage{booktabs}

\usepackage[pagebackref,breaklinks,colorlinks,citecolor=eccvblue]{hyperref}

 \usepackage{colortbl}   
\usepackage{tablefootnote}
\usepackage{booktabs}
\usepackage{xcolor}
\usepackage{multicol}
\usepackage{multirow}
\usepackage{pifont}

\usepackage{graphicx}

\definecolor{impgreen}{RGB}{0,120,90}

\usepackage{listings}
\usepackage{tcolorbox}
\tcbuselibrary{listings,breakable,skins}
\definecolor{myblue}{RGB}{218,232,252}
\lstdefinestyle{promptstyle}{
  basicstyle=\ttfamily\footnotesize,
  breaklines=true,
  breakatwhitespace=false,
  columns=fullflexible,
  keepspaces=true,
  showstringspaces=false,
  tabsize=2
}

\lstdefinestyle{promptstyle}{
  basicstyle=\ttfamily\footnotesize,
  breaklines=true,
  breakatwhitespace=false,
  columns=fullflexible,
  keepspaces=true,
  showstringspaces=false,
  tabsize=2
}

\newtcblisting{prompttemplate}[1]{
  enhanced,
  breakable,
  colback=gray!10,
  colframe=black,
  colbacktitle=white,
  coltitle=black,
  fonttitle=\bfseries,
  title={#1},
  center title,
  boxrule=0.5pt,
  arc=2pt,
  left=1mm,
  right=1mm,
  top=1mm,
  bottom=1mm,
  listing only,
  listing options={style=promptstyle}
}

\newtcolorbox{algobox}[1]{
  enhanced,
  breakable,
  colback=gray!6,
  colframe=black,
  colbacktitle=white,
  coltitle=black,
  fonttitle=\bfseries,
  title={#1},
  boxrule=0.5pt,
  arc=2pt,
  left=1mm,
  right=1mm,
  top=1mm,
  bottom=1mm
}

\newcounter{algorithmctr}
\renewcommand{\thealgorithmctr}{\arabic{algorithmctr}}

\definecolor{mygray}{RGB}{220,220,220}

\usepackage{multirow}
\usepackage{tabularx}
\usepackage{graphicx}
\usepackage{booktabs}

\definecolor{rone}{RGB}{30,90,180}
\definecolor{rtwo}{RGB}{20,130,95}
\definecolor{rthree}{RGB}{210,110,25}
\definecolor{posgreen}{RGB}{0,128,75}
\definecolor{negred}{RGB}{190,45,45}

\usepackage{float}

\begin{document}

\title{MT-EditFlow:  Reinforcement Learning for Multi-Turn  Image Editing with Flow Matching}

\titlerunning{Reinforcement Learning for Multi-Turn  Image Editing with Flow Matching}

\author{Jiahui Huang\inst{1}$^*$ \and
Yasi Zhang\inst{2}$^*$  \and
Tianyu Chen\inst{3}
\and
 Shu Wang \inst{1}
\and Jianwen Xie \inst{4} 
\and Oscar Leong\inst{2} \and Mingyuan Zhou\inst{3} \and Nanzhu Wang\inst{1} \and Ying Nian Wu\inst{2}
}

\authorrunning{J. Huang et al.}

\institute{Apple \and University of California, Los Angeles \and University of Texas at Austin \and Lambda, Inc \\ $^*$Equal Contribution.
}

\maketitle

\begin{abstract}

Recent breakthroughs in instruction-based image editing have captured significant attention, as models are now capable of handling real-world editing demands with the practicality required by everyday users. However, editing models trained primarily for single-turn edits often break down in multi-turn editing—the natural interactive setting where a user iteratively refines an image based on the model’s own previous outputs. This failure stems from the all-or-nothing requirement, where a single failed turn compromises the entire sequence, and error propagation, where exposure bias leads to compounding editing errors.
To address these challenges, we introduce MT-EditFlow, a flow-matching reinforcement learning framework designed to optimize reward signals for sequential image editing. MT-EditFlow integrates a multi-turn perspective with a multi-reward formulation to provide a unified structure applicable to both GRPO and NFT-based reinforcement learning methods. We systematically analyze and optimize the reward signal by investigating effective scoring strategies for turn-level aggregation, VLM reasoning modes to trade off reward bias and variance, and advantage fusion levels to prevent reward hacking. Our findings reveal that broadcasting the aggregated advantage across the entire editing trajectory effectively bridges the gap between local planning and global multi-turn task success.
Extensive experiments demonstrate that MT-EditFlow significantly improves performance across diverse base models. Notably, it boosts FLUX.1-Kontext-dev by 6.85 points in turn-3 overall performance, surpassing state-of-the-art open-source models such as Qwen-Image-Edit. By maintaining high marginal success rates and reducing exposure bias,  MT-EditFlow provides a foundation for more reliable and natural human-AI collaboration in visual content creation.

\keywords{Multi-turn Image Editing \and Flow Matching \and Reinforcement Learning }
\end{abstract}

\begin{figure}
    \centering
\includegraphics[width=1\linewidth]{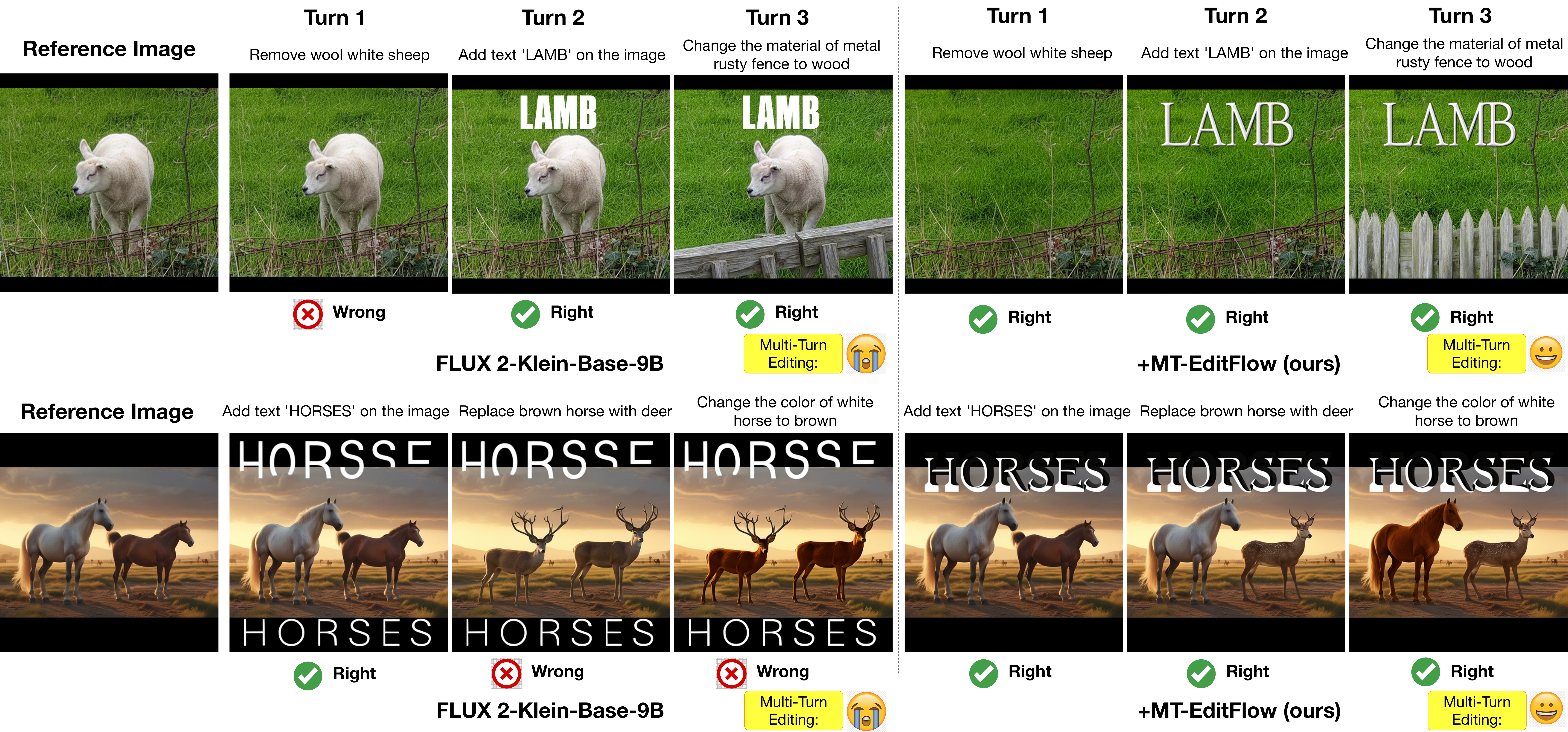}
    \caption{\textbf{Qualitative results of MT-EditFlow (Ours).} In multi-turn image editing, a single failed turn compromises the entire sequence, even if other turns succeed (e.g., a sheep failing to be removed). Furthermore, \textit{error accumulation} is a common pitfall: if an intermediate edit is incorrect, subsequent turns are often derailed. For instance, using the FLUX.2 base model, if a white horse is mistakenly transformed into a deer during the second turn, the third edit intended for the horse is effectively doomed. By contrast, MT-EditFlow enables the model to execute robust, successful edits across multiple consecutive turns, effectively preventing error propagation. }
    \label{fig:qualitative}
    \vspace{-0.5cm}
\end{figure}

\section{Introduction}
\label{sec:intro}

Recent advancements in instruction-based image editing \cite{comanici2025gemini25pushingfrontier, openai2025image4o, seedream, flux-2-2025, labs2025flux1kontextflowmatching, wu2025omnigen2, xiao2025omnigen, liu2025step1x, wu2025qwenimagetechnicalreport} have enabled models to handle complex, real-world requests. However, a significant gap remains: while humans naturally edit images through iterative, multi-turn dialogues \cite{ye2025imgedit, yang2025complexedit, edival_agent}, existing models are primarily trained for ``one-shot'' interactions. When applied to sequential editing, these models suffer severe degradation in instruction adherence and visual consistency. This degradation stems from two primary factors (see Fig. \ref{fig:qualitative} for qualitative illustrations):
\begin{itemize}
    \item \textbf{The ``All-or-Nothing'' Requirement:} Multi-turn editing demands sequential perfection. A single failed turn (e.g., a sleep failing to be removed) compromises the entire sequence, leaving the user dissatisfied even if all other edits succeed. 
    \item \textbf{Error Propagation:} Existing models struggle when conditioned on their own imperfect outputs, known as exposure bias \cite{schmidt2019generalization, edival_agent}. Errors compound over multiple steps, derailing subsequent turns. For instance, if the model mistakenly transforms a white horse into a deer in turn two, a third edit intended for the horse is effectively doomed.
\end{itemize}
Recent post-training methods for flow-matching models \cite{liu2022flow, lipman2022flow}, including GRPO \cite{flow_grpo, xue2025dancegrpo} and DiffusionNFT \cite{diffusion_nft}, demonstrate that reinforcement learning (RL) \cite{sutton1999reinforcement} can effectively align generative models \cite{rombach2022highstablediffusion, flux-2-2025, wu2025qwenimagetechnicalreport} with reward signals. Nevertheless, these approaches strictly target single-shot image generation. Because they operate on a single-turn, single-reward paradigm, they do not explicitly model the temporal dependencies of multi-turn interactions, nor do they address the quality drift that arises during chained edits. 

To address these challenges, we introduce \textbf{MT-EditFlow}, a flow-matching reinforcement learning framework that tailors reward signal design for sequential image editing by integrating a \textit{multi-turn} perspective with a \textit{multi-reward} formulation, ultimately providing a unified reward structure applicable to both GRPO- and NFT-based RL methods. We focus on algorithms based on flow-matching models due to their superior performance, scalability, and fast inference efficiency, as demonstrated by recent open-source SOTA editing models such as Qwen-Image-Edit \cite{wu2025qwenimagetechnicalreport}, the FLUX series \cite{flux-2-2025, labs2025flux1kontextflowmatching}, and Step1X-Edit \cite{liu2025step1x}.
Prior work implements editing evaluation by zero-shot prompting a SOTA VLM (usually closed-source) \cite{ye2025imgedit, liu2025step1x, li2025uniworldv2editr1}, utilizing a fine-tuned VLM for more specialized evaluation \cite{luo2026editscore, wu2026editreward}, or employing a combination of VLMs and expert tools in an agentic framework \cite{edival_agent}. For fast RL training, MT-EditFlow adopts zero-shot prompting of SOTA VLMs (i.e., Qwen3-VL-8B \cite{yang2025qwen3}) to evaluate Instruction Following (IF), paired with EdiVal-CC \cite{edival_agent} to precisely measure  Content Consistency (CC) in terms of semantic-level distance in unedited regions. Crucially, we comprehensively study and optimize reward signal design—an area notably underexplored in previous work—by systematically analyzing:
\begin{itemize}
    \item \textbf{Scoring Strategies:} How to effectively aggregate multiple turn-level rewards.
     \item \textbf{Evaluator Modes:} The trade-off between reward bias and variance when comparing ``thinking'' versus ``non-thinking'' VLM modes.
    \item \textbf{Advantage Fusion Levels:} How to balance competing reward aspects (IF vs. CC) to avoid reward hacking.
\end{itemize}
 
To support this framework, we constructed $\sim$2.5K high-quality multi-turn prompt chains with GPT-4o \cite{achiam2023gpt4} using the EdiVal-Agent pipeline \cite{edival_agent}. We then applied MT-EditFlow to extend both GRPO-based and NFT-based diffusion RL methods to the sequential editing setting. Our results demonstrate that broadcasting the aggregated advantage across the entire multi-turn trajectory effectively aligns local planning decisions with global task success, ensuring more cohesive sequential edits.
Extensive experiments demonstrate that MT-EditFlow significantly improves sequential editing performance across diverse base models. Notably, it boosts FLUX.1-Kontext-dev \cite{labs2025flux1kontextflowmatching} by 6.85 points in turn-3 overall performance, surpassing state-of-the-art open-source Qwen-Image-Edit \cite{wu2025qwenimagetechnicalreport} models, and improves FLUX.2-klein-base \cite{flux-2-2025} by 2.90 points, pushing the boundary of current open-source models. By maintaining high marginal editing success rates, MT-EditFlow consistently reduces exposure bias over successive turns. Ultimately, our results show that MT-EditFlow provides a unified reward signal design applicable to both GRPO and NFT-style RL algorithms, effectively unlocking robust, multi-turn capabilities in flow-based image editors.

\section{Related Work}
\subsubsection{Instruction-Based Image Editing}

InstructPix2Pix (IP2P) \cite{brooks2023instructpix2pix} proposed a two-stage procedure for turning a text-to-image diffusion model \cite{rombach2022highstablediffusion, zhang2024object} into an image editor: (i) generate paired editing examples using Stable Diffusion \cite{rombach2022highstablediffusion} together with training-free methods such as Prompt-to-Prompt \cite{hertz2023prompt2prompt}; and (ii) fine-tune the diffusion model on these synthetic pairs. Follow-up diffusion-based systems, including MagicBrush \cite{zhang2023magicbrush}, UltraEdit \cite{zhao2024ultraedit}, and AnyEdit \cite{yu2025anyedit}, extend this approach to large-scale, fine-grained editing of real images. More recent approaches (e.g., OmniGen \cite{xiao2025omnigen, wu2025omnigen2}, Step1X-Edit \cite{liu2025step1x}, FLUX.1 Kontext \cite{labs2025flux1kontextflowmatching}, Qwen-Image-Edit \cite{wu2025qwenimagetechnicalreport}, and Seedream \cite{seedream}) introduce task-specialized architectures and increasingly adopt flow-matching frameworks \cite{liu2022flow, lipman2022flow, zhang2024flow}.
In parallel, a separate research direction has focused on LLM-native editors (typically closed-source), such as Gemini~2.0 Flash Image \cite{google2025gemini2}, Nano Banana 1/2 \cite{comanici2025gemini25pushingfrontier}, and GPT-Image-1/1.5 \cite{openai2025image4o}. These systems support  {in-context multi-turn editing}: users iteratively adjust an image through dialogue, while the model tracks and leverages a consistent editing history—an interaction mode shown to be   effective for multi-step image refinement.

To the best of our knowledge, we present the first RL-based framework aimed at narrowing the gap between current open-source editors and such multi-turn editing capabilities, with a particular focus on a unified reward signal design.

\subsubsection{Image Editing Evaluation and Benchmarks}

Recent work relies exclusively on vision-language models (VLMs) as interpretable judges—e.g., VIEScore \cite{ku2023viescore}, GEdit-Bench \cite{liu2025step1x}, I$^2$EBench \cite{ma2024i2ebench}, HQ-Edit \cite{hui2024hqedit}, Complex-Edit \cite{yang2025complexedit}, and ImgEdit \cite{ye2025imgedit}—by querying models such as GPT-4o \cite{openai2025image4o} about specific aspects of an edit. While VLMs provide holistic, language-mediated assessments, they fall short in several ways: they are poor at spatial reasoning \cite{zhang2025dospatial, cheng2024spatialrgpt, chen2024spatialvlm,qharabagh2024lvlm, chang2025skews}, prone to hallucinating objects and attributes \cite{bai2024hallucination}, insensitive to subtle pixel-level changes \cite{vo2025visionbiased} (e.g., fine structures or small attribute shifts), and miscalibrated for artifacts and aesthetics \cite{richhf, xu2023imagereward, hpsv3}. EdiVal-Agent \cite{edival_agent} mitigates these limitations by integrating VLM reasoning with grounding tools, symbolic and object-centric pixel- and semantic-level metrics, and human preference models, producing precise and interpretable evaluation for instruction-based editing.


In this paper, we evaluate MT-EditFlow primarily on the multi-turn benchmark EdiVal-Bench \cite{edival_agent}, with supplementary evaluations on widely-used single-turn benchmarks, ImgEdit \cite{ye2025imgedit}.


\subsubsection{Reinfocement Learning with Diffusion and Flow Matching}

A central challenge in extending RL \cite{sutton1999reinforcement} from discrete autoregressive models \cite{gpt3, achiam2023gpt4} to diffusion models \cite{ho2020denoising, song2020score} lies in the intractability of exact likelihoods in diffusion \cite{song2021maximum}, which undermines many standard RL algorithms. Early PPO-style \cite{schulman2017proximal} adaptations, such as DDPO \cite{black2023trainingddpo} and DPOK \cite{fan2023dpok}, factorize trajectory likelihoods step-by-step along the reverse process, but ignore forward consistency. Subsequent GRPO-based \cite{guo2025deepseek} extensions, including Flow-GRPO \cite{flow_grpo} and DanceGRPO \cite{xue2025dancegrpo}, discretize reverse diffusion to enable gradient optimization and demonstrate improved scalability. However, they tightly couple the training objective with specific SDE solvers, leading to forward–reverse inconsistency, solver restrictions, and limited efficiency. MixGRPO \cite{li2025mixgrpounlockingflowbasedgrpo} enhances efficiency by blending SDE and ODE formulations, yet retains these structural constraints. More recently, Diffusion Negative-aware FineTuning (DiffusionNFT) \cite{diffusion_nft} reframes diffusion RL through the forward process via flow matching, contrasting positive and negative samples to define implicit policy improvement without explicit likelihood estimation or reliance on particular samplers. This decoupled formulation substantially improves efficiency over Flow-GRPO and even eliminates the need for classifier-free guidance \cite{ho2022classifiercfg}.

In our work, MT-EditFlow extends both Flow-GRPO and DiffusionNFT with MT-Reward to enable effective RL for multi-turn image editing, and demonstrates consistent improvements across different base models.


\section{Background}


\subsection{Flow Matching}
Flow Matching (FM) \cite{liu2022flow, lipman2022flow} defines a continuous-time transformation between a noise distribution $p_1$ and a data distribution $p_0$. Let ${x}_0 \sim p_0$ be a data sample and ${x}_1 \sim p_1$ be standard Gaussian noise. A probability path $p_t({x})$ is constructed to bridge these distributions. In the case of \textit{Rectified Flow} \cite{liu2022flow, simeoni2025dinov3}, the latent $x_t$ is defined by the   linear interpolation: $
  {x}_t = (1-t){x}_0 + t {x}_1.
$
The model is trained to learn a time-dependent vector field $v_\theta({x}_t, t)$ that predicts the velocity $v_t = \frac{d{x}_t}{dt} = {x}_1 -{x}_0$. The flow matching objective is minimized as:
\begin{equation}
    \mathcal{L}_{FM}(\theta) = \mathbb{E}_{t \sim \mathcal{U}[0,1], {x}_0 \sim p_0, {x}_1 \sim p_1} \left[ \|v_\theta({x}_t, t) - ({x}_1 - {x}_0)\|^2 \right].
\end{equation}
After training is done, inference is usually performed by numerically solving the ODE $d{x}_t = v_\theta({x}_t, t)dt$ from $t=1$ to $t=0$ using an ODE solver.

\subsection{Reinforcement Learning  for Image Editing}

Let $\mathcal{D}$ denote the training set, $(x_f,c)$ be the pair of  reference image and editing instruction in the training dataset, and $\pi_\theta$ is the policy induced by the flow matching editing model. Fundamentally, the objective of reinforcement learning \cite{sutton1999reinforcement} for single-turn image editing is to maximize the reward \cite{li2025uniworldv2editr1, luo2026editscore}:
\begin{align}
   \mathcal{J}({\theta}) = \mathbb{E}_{(x_f,c)\sim \mathcal{D},  x_0 \sim \pi_{\theta}(\cdot|x_f,c)}  \left[ r (x_0,x_f, c) \right], 
\end{align}
where the reward function $r$ evaluates editing success. While $r$ may comprise evaluations across multiple dimensions of an edit, we aggregate them into a single, unified function $r$ for conciseness.

Extending Reinforcement Learning (RL) \cite{sutton1999reinforcement} from discrete autoregressive models \cite{gpt3, achiam2023gpt4} to diffusion and flow models \cite{ho2020denoising, liu2022flow} is challenging due to the intractability of exact likelihoods. 

\subsubsection{Flow-GRPO} Flow-GRPO \cite{flow_grpo} addresses this by approximating the likelihood via a stochastic denoising process. Using an SDE sampler \cite{song2020score} that shares the same marginal distribution as the deterministic flow, Flow-GRPO enables gradient-based optimization through Euler-Maruyama discretization, where each sampling step is modeled as a Gaussian distribution.
 For a given reference image $x_f$ and editing instruction $c$, the policy $\pi_\theta$ generates a group of $G$ outputs $\{x_{0, i}\}_{i=1}^G$. Each sample $x_{0, i}$ is produced through a $T$-step stochastic process $\{x_{t, i}\}$ starting from Gaussian noise $x_{1, i} \sim \mathcal{N}(\mathbf{0}, \mathbf{I})$ and terminating at data $x_{0, i}$. The advantage $A_i$ is computed by normalizing the rewards $R_i = r(x_{0, i}, x_f, c)$ within the group:
\begin{equation}\label{eq:adv}
    A_i := \text{Adv}(R_i) = \frac{R_i - \frac{1}{G}\sum_{j=1}^G R_j}{\text{std}(\{R_j\}_{j=1}^G)}.
\end{equation}
The Flow-GRPO objective is optimized by maximizing (we ignore the expectation condition for brevity):
\begin{equation}
    \mathcal{J}_{\text{Flow-GRPO}}(\theta) = \mathbb{E} \left[ \frac{1}{G} \sum_{i=1}^G \left( \frac{1}{T} \sum_{t=1}^{T} \left(\mathcal{L}_{\text{clip}}^{\text{GRPO}}(\rho_{t, i}(\theta), A_i) - \beta \mathbb{D}_{KL}(\pi_\theta \| \pi_{\text{ref}})\right) \right) \right], \label{eq:editflow_obj}
\end{equation}
where the clipped surrogate  
$\mathcal{L}_{\text{clip}}^{\text{GRPO}}(\rho, A) = \min \left( \rho A, \text{clip}(\rho, 1-\epsilon, 1+\epsilon) A \right),
$ the step-wise importance ratio 
    $\rho_{t, i}(\theta) = \frac{p_\theta(x_{t-\Delta t, i} | x_{t, i}, x_f, c)}{p_{\theta_{\text{old}}}(x_{t-\Delta t, i} | x_{t, i}, x_f, c)},$ and the $\pi_{\text{ref}}$ is the reference policy used for regularization.

\subsubsection{DiffusionNFT} 
Another line of work, DiffusionNFT \cite{diffusion_nft}, performs implicit policy improvement by contrasting positive and negative samples, without explicit likelihood estimation or reliance on specific samplers. Specifically, the DiffusionNFT objective is defined as (we ignore the expectation condition for brevity):
\begin{align}
&\mathcal{J}_{\text{DiffusionNFT}}(\theta) = \notag \\
&\mathbb{E}\left[  \frac{1}{G} \sum_{i=1}^G \left( \frac{1}{T} \sum_{t=1}^{T} \left(\mathcal{L}_{\text{clip}}^{\text{NFT}}(A_i) \| v^+_\theta  - v \|^2_2 + (1 - \mathcal{L}_{\text{clip}}^{\text{NFT}}(A_i)) \| v^-_\theta  - v \|^2_2\right)  \right) \right].
\end{align}
Here $v^+_\theta(x_{t,i}, t)$ and $v^-_\theta(x_{t,i}, t)$ denote the positive and negative velocities parameterized by $\theta$ with inputs $(x_{t,i}, t, x_f, c)$, and $A_i$ is computed in the same way as in Eq. \ref{eq:adv}. In practice, the clipped reward function is defined as
$
\mathcal{L}_{\text{clip}}^{\text{NFT}}(A):= \frac{1}{2} + \frac{1}{2} \text{clip}\left(A/z_{\max}, -1, 1\right).$

\section{MT-EditFlow:  RL for Multi-Turn Image Editing}

We note that existing RL methods for flow-matching models typically consider only a single reward (excluding the KL divergence with a reference model) and operate in a single-turn setting. In contrast, MT-EditFlow extends this framework from two perspectives: a \textit{multi-reward} structure and a \textit{multi-turn} formulation. Specifically, we introduce rewards for both instruction following (IF) and content consistency (CC), which are the two most important aspects of image editing.
Our goal is to provide a unified reward signal design for multi-turn editing that is applicable to both GRPO-style and NFT-style RL algorithms.

We first formulate the multi-turn editing setting. We then identify three key factors that significantly affect the reward signal's effectiveness. In particular, we study: (1) the score aggregation strategy across turns for effective RL optimization, (2) the fusion level of multiple rewards to avoid reward hacking, and (3) the use of thinking-mode reward evaluation to trade off reward bias and variance and boost overall performance.

\subsection{MT-EditFlow Reward Objective}

Suppose we have a multi-turn editing task with $K$ turns. The training dataset $\mathcal{D}$ consists of pairs $(x_f, \mathcal{C})$, where $\mathcal{C} = [c_1, c_2, \ldots, c_K]$ is a sequence of editing instructions applied to the same reference image $x_f$ in order. As a result, the model produces a sequence of edited images $\mathcal{X} = [x_0^{(1)}, x_0^{(2)}, \ldots, x_0^{(K)}]$. 
We denote $\mathcal{X}_{\text{int}} = [x_0^{(1)}, x_0^{(2)}, \ldots, x_0^{(K-1)}]$ as the intermediate edits, which are typically not directly evaluated by users. Strictly speaking, the multi-turn editing task can be defined solely based on the final result $(x_0^{(K)}, x_f, \mathcal{C})$, without explicitly considering the intermediate edits, as illustrated in Fig.~\ref{fig:qualitative}. 
In the following, we first discuss two key aspects of editing quality at each turn, and then describe how these two aspects are fused at the reward level.

\subsubsection{Instruction Following (IF)}Marginal instruction-following success at turn $k$ is defined over $(x_0^{(k)}, x_0^{(k-1)}, c_k)$. For notation consistency, we define $x_0^{(0)} := x_f$. This follows the standard single-turn editing formulation.
Prior work evaluates editing performance by zero-shot prompting a state-of-the-art (SOTA) VLM (often closed-source) \cite{ye2025imgedit, liu2025step1x, li2025uniworldv2editr1}, using fine-tuned VLMs for more specialized evaluation \cite{luo2026editscore, wu2026editreward}, or employing a combination of VLMs and expert tools in an agentic framework \cite{edival_agent}. In this work, for RL training efficiency and simple reward evaluation, we implement the instruction-following (IF) reward by prompting the open-source VLM Qwen3-VL-8B. 
Specifically, we query a VLM reward model with the inputs $\{(x_0^{(k)}, x_0^{(k-1)}, c_k)\}_{k=1}^K$ and obtain the IF score $r_{\mathrm{IF}}^{(k)} = \Phi(x_0^{(k)}, x_0^{(k-1)}, c_k)$ for each turn $k$, where $\Phi$ denotes the VLM evaluator.
Under the strict multi-turn definition, overall success can be defined as $r_{\mathrm{IF}}^{\mathrm{mt}} = \bigwedge_{k=1}^{K} r_{\mathrm{IF}}^{(k)}$, where $\bigwedge$ denotes the logical AND operator. However, in our experiments we find that this strict formulation does not provide an effective learning signal for RL. We therefore also consider an alternative aggregation that averages the turn-level rewards: $r_{\mathrm{IF}}^{\mathrm{avg}} = \frac{1}{K} \sum_{k=1}^K r_{\mathrm{IF}}^{(k)}$.

\subsubsection{Content Consistency (CC)} Prior work typically computes pixel-level or semantic distances directly between the reference image and the edited image \cite{zhang2023magicbrush, zhao2024ultraedit, yu2025anyedit, sheynin2024emuedit}. This approach ignores the fact that regions intended to be changed should not be included in the consistency evaluation. To more precisely measure content consistency (CC) and capture the essence of a successful edit, we adopt EdiVal-CC, which is proposed in \cite{edival_agent}. 
Formally, we denote our CC reward as $r_{\mathrm{CC}}^{(k)} = \text{EdiVal-CC}(x_0^{(k)}, x_0^{(0)}, \Omega)$, where $\Omega$ is provided by EdiVal-CC and indicates the regions that should remain unchanged, excluding areas intended for editing. Unlike instruction-following, $r_{\mathrm{CC}}^{(k)}$ already aggregates consistency across the editing chain by comparing the reference image with the $k$-turn image. Note that evaluating consistency between the $(k-1)$-turn and $k$-turn images is not meaningful, as the $(k-1)$-turn may already have deviated significantly from the reference image even if the $k$-turn appears consistent.

 \subsubsection{Discussion on Visual Quality (VQ)} As discussed in GIE-Bench \cite{gie_bench} and EdiVal-Bench \cite{edival_agent}, editing models such as GPT-Image-1 that prioritize visual quality often exhibit poor fidelity to the original image’s content and style. Consequently, using a pure VQ reward, or computing VQ differences between the reference and edited images, remains inconclusive. Therefore, we do not include VQ in our reward structure.
 
\subsubsection{Reward-Level Fusion} Straightforwardly, we obtain the final reward by a weighted sum of the IF and CC rewards (at index $i$ within the group), extending the single-turn reward $R_i$ to the multi-turn setting: 
\begin{equation}
R_i^{\textbf{MT-EditFlow}}(\mathcal{X}_i, x_f, \mathcal{C}) = \lambda_{\mathrm{IF}} r_{i,\mathrm{IF}}^{\mathrm{}} + \lambda_{\mathrm{CC}} r_{i,\mathrm{CC}}^{(K)},
\label{eq:reward_fusion}
\end{equation}
where $r_{i,\mathrm{IF}}^{\mathrm{}}$ has 3 different variants as described below. We fix $\lambda_{\mathrm{IF}} = 1$ and sweep $\lambda_{\mathrm{CC}}$ in our main experiments.   

\subsection{Scoring Strategy: Granularity and Sparsity}We prompt the VLM with different scoring rubrics. Specifically, we consider two levels of granularity: a binary success indicator $y_i^{(k)} \in \{0,1\}$ for turn $k$ of sample $i$, and a fine-grained Likert score $s_i^{(k)} \in \{1,\dots,5\}$.
We further study different operators for aggregating turn-level scores across the trajectory. Combining both dimensions, we consider the following three scoring strategies:
\begin{itemize}
    \item \textbf{Binary MT ($r_{i,\mathrm{IF}}^{\mathrm{mt}}$):} 
    A strict indicator where
    $r_{i,\mathrm{IF}}^{\mathrm{mt}} = \bigwedge_{k=1}^{K} y_i^{(k)}$,
    yielding $1$ only if all turns succeed.

    \item \textbf{Binary Average ($r_{i,\mathrm{IF}}^{\mathrm{avg}}$):}
    The mean of per-turn binary success,
    $r_{i,\mathrm{IF}}^{\mathrm{avg}} = \frac{1}{K}\sum_{k=1}^{K} y_i^{(k)}$.

    \item \textbf{Fine-grained Average ($r_{i,\mathrm{IF}}^{\mathrm{fg}}$):}
    The  mean of fine-grained scores,
    $r_{i,\mathrm{IF}}^{\mathrm{fg}} = \frac{1}{K}\sum_{k=1}^{K} s_i^{(k)}$.
\end{itemize}
While $r_{i,\mathrm{IF}}^{\mathrm{mt}}$ aligns with the strict definition of multi-turn success, it suffers from extreme reward sparsity. For a per-turn success probability $p$, the expected number of positive samples in a group of size $G$ is $G p^K$. As $K$ increases, the probability of a null gradient or high-variance advantage increases. In contrast, the fine-grained reward $r_{i,\mathrm{IF}}^{\mathrm{fg}}$ provides partial credit, preserving a dense learning signal even when most of turns fail.

\begin{figure}[t]
    \centering
    \includegraphics[width=0.98\linewidth]{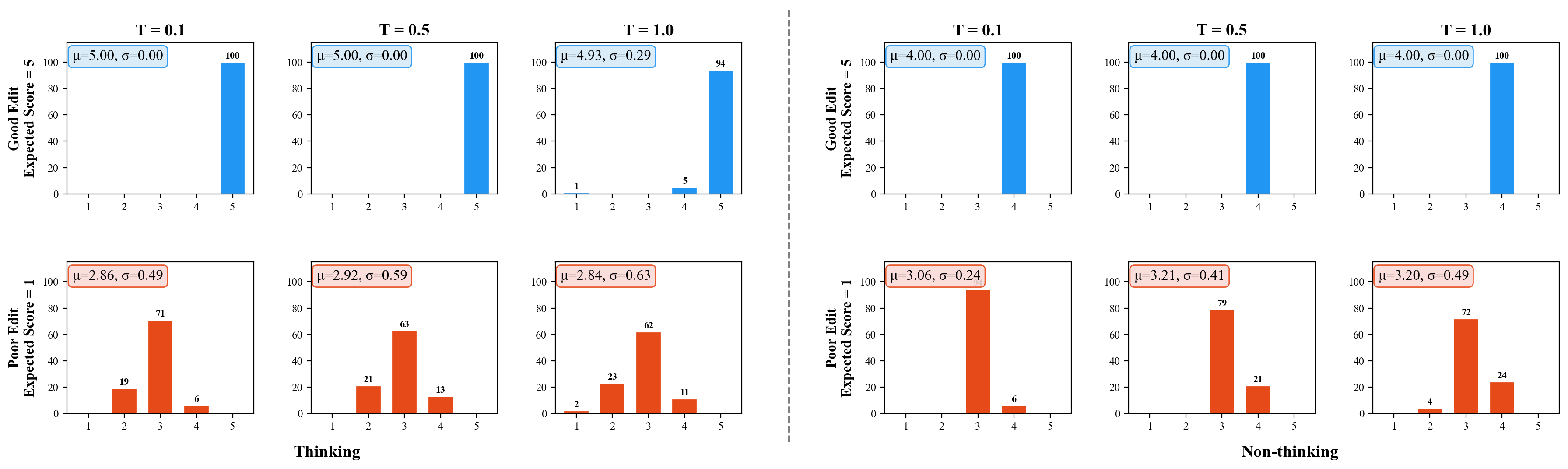}
    \caption{\textbf{
  Illustration of the reward bias and variance trade-off between thinking and non-thinking modes.}    Thinking mode consistently leads to reduced reward bias and increased variance across different temperatures. For example, the mean evaluation score under thinking mode for poor edits is closer to 1, the lowest score.}
    \label{fig:score_distribution}
    \vspace{-0.5cm}
\end{figure}

\subsection{Evaluator Mode: Thinking vs. Non-thinking}
The reliability of the VLM reward model $\Phi$ is influenced by the prompting structure. We compare a \textbf{Non-thinking} mode (direct score output) with a \textbf{Thinking} mode, where the VLM is prompted to generate a Chain-of-Thought (CoT) analysis of the editing trajectory $\mathcal{X}$ relative to $x_f$ and $\mathcal{C}$ before producing $r_{\mathrm{IF}}^{(k)}$. 
Empirically, the thinking mode is better aligned with human judgments: the non-thinking mode can mis-score obvious good edits and occasionally assign overly high scores to poor edits, while the thinking mode produces a more accurate score distribution, albeit with higher variance (Fig.~\ref{fig:score_distribution}). We provide the prompt details in the Appendix.

\subsection{Fusion Level: Reward vs. Advantage}
When jointly optimizing Instruction Following (IF) and Content Consistency (CC), we compare two fusion levels. \textbf{Reward-Level Fusion} (Eq.~\ref{eq:reward_fusion}) sums the raw signals before computing a single group-relative advantage $A_i$. However, because $r_{\text{IF}}$ and $r_{\text{CC}}$ operate on different scales and noise distributions, this can hurt the policy optimization. 
Alternatively, we propose \textbf{Advantage-Level Fusion}, where we compute independent advantages $A_{i, \text{IF}}$ and $A_{i, \text{CC}}$ for each objective and then aggregate them:
\begin{equation}\label{eq:adv_fusion}
    A_i^{\text{MT-EditFlow}} = \lambda_{\text{IF}}A_{i, \text{IF}} + \lambda_{\text{CC}} A_{i, \text{CC}},
\end{equation}
where $A_{i, \text{IF}} := \text{Adv}(R_{i, \text{IF}}) = \text{Adv}(r_{i, \text{IF}}(\mathcal{X}_i, x_f, \mathcal{C}))$, and similarly for $A_{i, \text{CC}}$.
By standardizing the signals within the group before fusion, we prevent one reward component from dominating the gradient direction due to scale imbalances.

\subsection{Trajectory-Level Advantage Broadcasting}

Given the multi-turn reward defined in Eqs.  \ref{eq:reward_fusion} and \ref{eq:adv_fusion}, 
we broadcast the same trajectory-level advantage to every turn in the editing chain:
\begin{align}
    A_i^{(k)} :=  \text{Adv}(R_i^{\textbf{MT-EditFlow}})\quad\text{or} \quad A_i^{\textbf{MT-EditFlow}} , \quad k = 1,\dots,K .
\end{align}
For both Flow-GRPO and DiffusionNFT, this mechanism propagates a single trajectory-level reward signal to all denoising steps and editing turns. As a result, the optimization objective aligns local planning decisions with the global success of the multi-turn editing trajectory.

\begin{table}[ht]
\centering
\caption{\textbf{Results of multi-turn editing on EdiVal-Bench.} 
EdiVal-IF (Instruction Following), EdiVal-CC (Content Consistency), and EdiVal-O (Overall) across three sequential editing turns.}
\label{tab:overall}
{
\begin{tabular}{l ccc ccc | ccc}
\toprule
& \multicolumn{3}{c}{\textbf{EdiVal-IF}} 
& \multicolumn{3}{c}{\textbf{EdiVal-CC}} 
& \multicolumn{3}{c}{\textbf{EdiVal-O}} \\
\cmidrule(lr){2-4} \cmidrule(lr){5-7} \cmidrule(lr){8-10}
\textbf{Model} & T1 & T2 & T3 & T1 & T2 & T3 & T1 & T2 & T3 \\
\midrule

\rowcolor{mygray}
\multicolumn{10}{c}{\textbf{\textit{Closed-Source Models}}} \\

Seedream 4.0        & 75.93 & 55.58 & 41.59 & 92.51 & 88.03 & 85.86 & 83.81 & 69.95 & 59.76 \\
GPT-Image-1.5       & 75.19 & 55.92 & 40.08 & 94.49 & 91.20 & 88.49 & 84.29 & 71.41 & 59.55 \\
Nano Banana 2       & 73.89 & 54.17 & 38.61 & 93.54 & 90.52 & 88.61 & 83.14 & 70.02 & 58.49 \\
FLUX.2-max          & 75.55 & 55.27 & 39.36 & 92.91 & 88.30 & 85.78 & 83.78 & 69.86 & 58.10 \\
Nano Banana         & 70.70 & 50.66 & 35.35 & 93.91 & 90.48 & 89.48 & 81.48 & 67.70 & 56.24 \\
GPT-Image-1         & 73.12 & 54.89 & 38.35 & 81.00 & 77.78 & 75.50 & 76.96 & 65.34 & 53.81 \\
FLUX.1-Kontext-max  & 69.49 & 46.89 & 31.83 & 93.93 & 90.90 & 88.40 & 80.79 & 65.29 & 53.04 \\
Gemini 2.0 Flash    & 68.07 & 45.96 & 28.42 & 90.58 & 85.10 & 80.88 & 78.52 & 62.54 & 47.94 \\

\midrule

\rowcolor{mygray}
\multicolumn{10}{c}{\textbf{\textit{Open-Source  Models}}} \\

IP2P            & 37.41 & 10.66 &  2.80 & 76.85 & 68.36 & 60.30 & 53.62 & 26.99 & 12.99 \\
MagicBrush      & 42.31 & 15.73 &  4.90 & 86.96 & 81.26 & 76.86 & 60.66 & 35.75 & 19.41 \\
AnyEdit         & 41.07 & 16.32 &  7.22 & 86.42 & 78.91 & 70.10 & 59.58 & 35.89 & 22.50 \\
UltraEdit       & 51.37 & 17.70 &  6.36 & 86.80 & 84.50 & 82.40 & 66.78 & 38.67 & 22.89 \\
OmniGen         & 54.72 & 24.48 & 10.66 & 93.00 & 88.42 & 83.92 & 71.34 & 46.52 & 29.91 \\
Step1X-Edit     & 61.89 & 34.97 & 17.83 & 92.76 & 88.52 & 85.21 & 75.77 & 55.64 & 38.98 \\
Qwen-Image-Edit & 72.90 & 44.06 & 22.55 & 84.22 & 80.52 & 77.98 & 78.36 & 59.56 & 41.93 \\
\midrule
 FLUX.1-Kontext-dev  & 59.97 & 32.69 & 16.61 &  {95.32} &  {92.24} &  {90.22} & 75.61 & 54.91 & 38.71 \\
\rowcolor{myblue}   + MT-EditFlow-NFT & \textbf{62.24} & \textbf{38.99} & \textbf{22.90} & \textbf{95.75} & \textbf{92.47} & \textbf{90.62} & \textbf{77.20} & \textbf{60.04} & \textbf{45.56}\\                     
   $\Delta$                                  
 & \textcolor{impgreen}{+2.27} 
& \textcolor{impgreen}{+6.30} 
& \textcolor{impgreen}{+6.29}     
& \textcolor{impgreen}{+0.43} 
& \textcolor{impgreen}{+0.23} 
& \textcolor{impgreen}{+0.40}       
& \textcolor{impgreen}{+1.59} 
& \textcolor{impgreen}{+5.13} 
& \textcolor{impgreen}{+6.85} \\                 
\midrule
  FLUX.2-Klein-base-9B & 67.83 & 45.63 & 28.67 & 94.57 & 90.52 & 87.25 & 80.09 & 64.27 & 50.01 \\

 \rowcolor{myblue}  + MT-EditFlow-NFT   & \textbf{70.28} & \textbf{48.08} & \textbf{31.47} & \textbf{95.18} & \textbf{91.53} & \textbf{88.95} & \textbf{81.79} & \textbf{66.34} & \textbf{52.91}\\
  $\Delta$
& \textcolor{impgreen}{+2.45} 
& \textcolor{impgreen}{+2.45} 
& \textcolor{impgreen}{+2.80}
& \textcolor{impgreen}{+0.61} 
& \textcolor{impgreen}{+1.01} 
& \textcolor{impgreen}{+1.70}
& \textcolor{impgreen}{+1.70} 
& \textcolor{impgreen}{+2.07} 
& \textcolor{impgreen}{+2.90}  
\\
\bottomrule
\end{tabular}
}
\vspace{-1cm}
\end{table}

\section{Experiments}

Our training and evaluation data are built on the Pico-Banana-400K dataset~\cite{qian2025picobanana400klargescaledatasettextguided}, whose images originate from OpenImages~\cite{kuznetsova2020openimages}. To construct multi-turn RL trajectories, we use the EdiVal-Agent pipeline~\cite{edival_agent} to generate object-centric and context-consistent instruction sequences with $K{=}3$ turns per reference image. 
In total, the dataset contains {2,319 reference images} with planned {3-turn} editing trajectories across 9 editing types and 12 semantic object categories, yielding \textbf{6,957 turn-level steps} for online RL fine-tuning.

For evaluation, we mainly utilize   {EdiVal-Bench} \cite{edival_agent}, which contains completely \textit{different} reference images and instruction chains from training to assess the performance of our model in multi-turn editing setting, with supplementary evaluations on widely-used single-turn benchmarks, ImgEdit \cite{ye2025imgedit}.

\subsubsection{Training Details}
We use FLUX.1-Kontext-dev \cite{labs2025flux1kontextflowmatching} and FLUX.2-klein-base-9B \cite{flux-2-2025} as our primary base models and fine-tune them with Low-Rank Adaptation (LoRA)~\cite{hu2022lora}. For efficiency, we employ Fully Sharded Data Parallelism (FSDP)~\cite{zhao2023pytorch} and DeepSpeed ZeRO~\cite{rajbhandari2020zero} alongside gradient checkpointing, and serve VLM-based rewards via a vLLM endpoint during online rollouts.

We train MT-EditFlow combined with either Flow-GRPO~\cite{flow_grpo} or DiffusionNFT~\cite{diffusion_nft} depending on the setting. Unless otherwise specified, we keep the training recipe fixed within each backbone/objective combination and vary only the factor of interest; we study reward signal design in Sec.~\ref{sec:ablation_reward} and provide other ablations in Sec.~\ref{sec:ablation}. We fix $\lambda_{\mathrm{IF}}{=}1$ and sweep $\lambda_{\mathrm{CC}}$ where indicated.

\begin{figure}[h]
\centering
\begin{minipage}[ht]{0.49\linewidth}
    \centering
    \includegraphics[width=\linewidth]{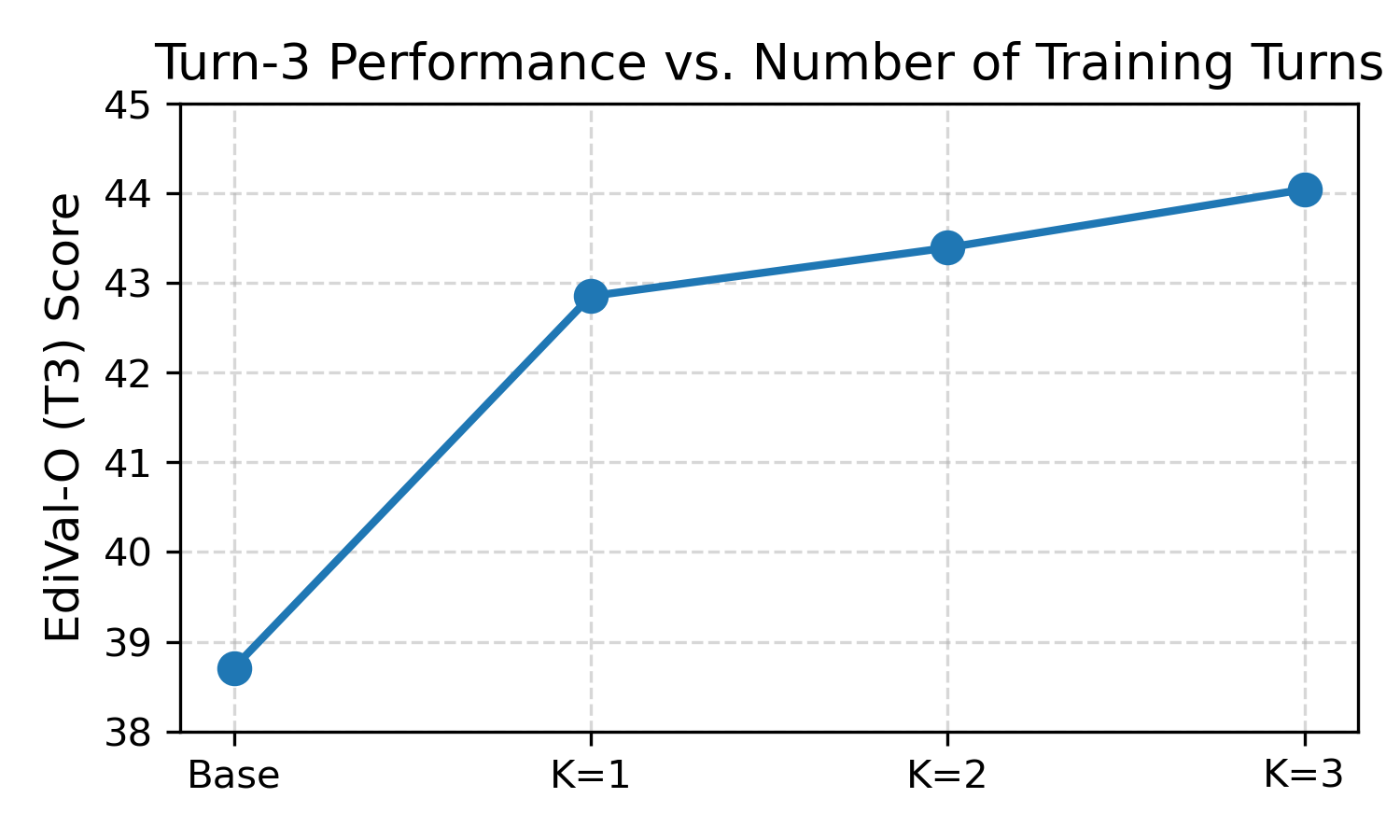}
    \captionof{figure}{Multi-turn editing performance increases with the number of training turns $K$ used.}
    \label{fig:t3}
\end{minipage}
\hfill
\begin{minipage}[ht]{0.47\linewidth}
    \centering
    \includegraphics[width=\linewidth]{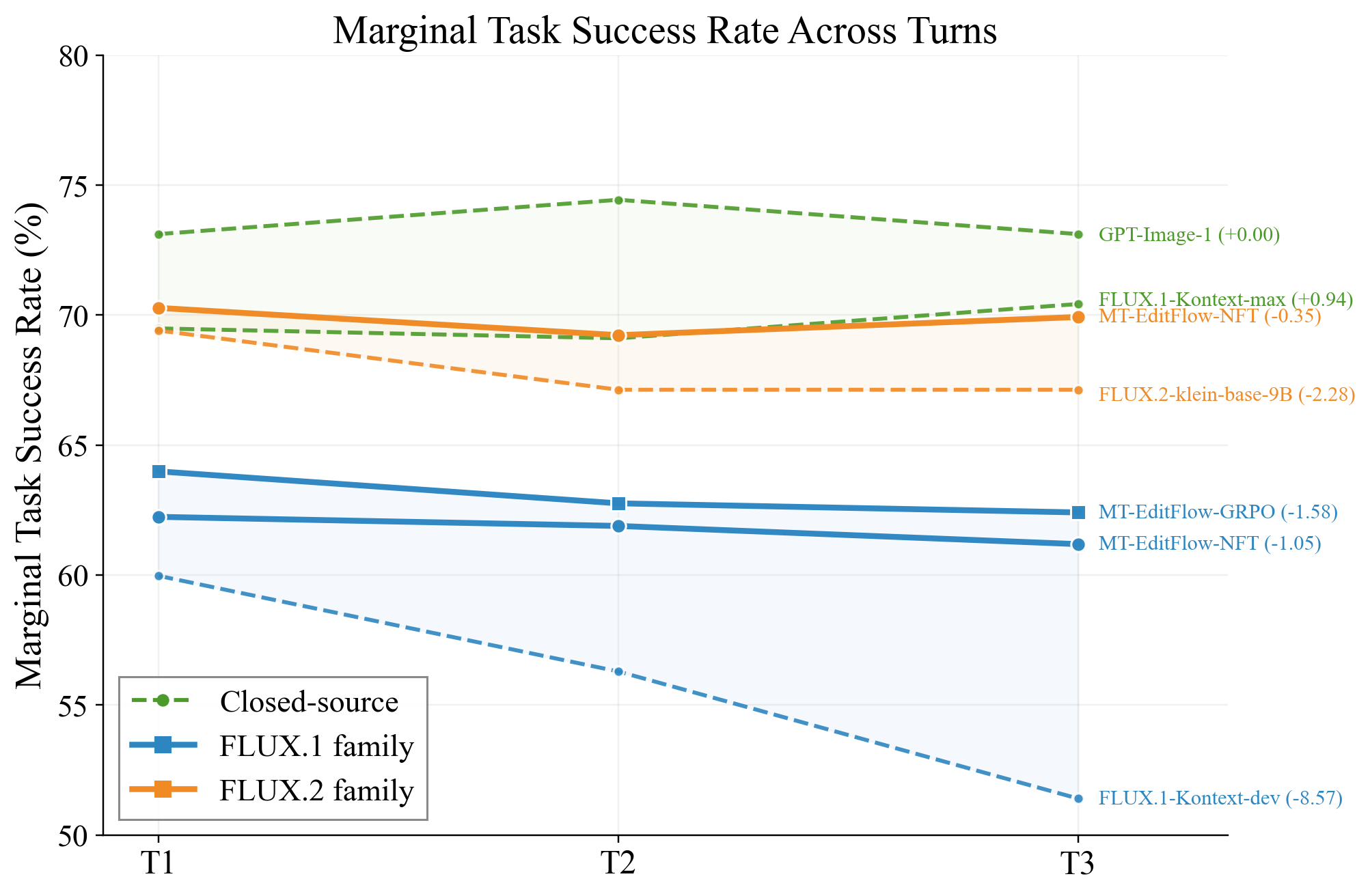}
    \captionof{figure}{Marginal task success across turns. With MT-EditFlow, performance remains stable, reducing exposure bias.}
    \label{fig:task_rate}
\end{minipage}
\vspace{-0.7cm}
\end{figure}

\begin{table}[ht]
\centering
\caption{Single-turn editing results of MT-EditFlow on ImgEdit-Bench. GPT-4.1 is used for evaluation. Our multi-turn RL algorithm naturally leads to improved single-turn performance.}
\label{tab:imgedit-bench}
\resizebox{1\linewidth}{!}
{
\begin{tabular}{l|ccccccccc|c}
\toprule
\textbf{Model} & \textbf{Add} & \textbf{Adjust} & \textbf{Extract} & \textbf{Replace} & \textbf{Remove} & \textbf{Background} & \textbf{Style} & \textbf{Hybrid} & \textbf{Action} & \textbf{Overall$\uparrow$} \\ \midrule
ICEdit        & 3.58 & 3.39 & 1.73 & 3.15 & 2.93 & 3.08 & 3.84 & 2.04 & 3.68 & 3.05 \\
Step1X-Edit   & 3.88 & 3.14 & 1.76 & 3.40 & 2.41 & 3.16 & 4.63 & 2.64 & 2.52 & 3.06 \\
UniWorld-V1   & 3.82 & 3.64 & 2.27 & 3.47 & 3.24 & 2.99 & 4.21 & 2.96 & 2.74 & 3.26 \\
BAGEL         & 3.81 & 3.59 & 1.58 & 3.85 & 3.16 & 3.39 & 4.51 & 2.67 & 4.25 & 3.42 \\
OmniGen2      & 3.57 & 3.06 & 1.77 & 3.74 & 3.20 & 3.57 & 4.81 & 2.52 & 4.68 & 3.44 \\
GPT-Image-1  & 4.61 & 4.33 & 2.90 & 4.35 & 3.66 & 4.57 & 4.93 & 3.96 & 4.89 & 4.20 \\
UniWorld-V2   & 4.29 & 4.44 & 4.32 & 4.69 & 4.72 & 4.41 & 4.91 & 3.83 & 4.83 & 4.49 \\ 
\midrule
FLUX.2-Klein-base-9B  & 4.42 & 4.23 & 2.30 & 4.42 & 3.86 & 4.23 & 4.94 & 3.43 & 4.72 & 4.06 \\
\rowcolor{myblue}{} + MT-EditFlow-NFT & 4.42 & 4.23 & 2.33 & 4.50 & 4.04 & 4.39 & 4.94 & 3.62 & 4.81 & 4.14 \\
$\Delta$ & -- & -- & \textcolor{impgreen}{+0.03} & \textcolor{impgreen}{+0.08} & \textcolor{impgreen}{+0.18} & \textcolor{impgreen}{+0.16} & -- & \textcolor{impgreen}{+0.19} & \textcolor{impgreen}{+0.09} & \textcolor{impgreen}{+0.08} \\
\bottomrule
\end{tabular}}
\end{table}

\begin{figure*}[t]
    \centering
    \includegraphics[width=1.0\linewidth]{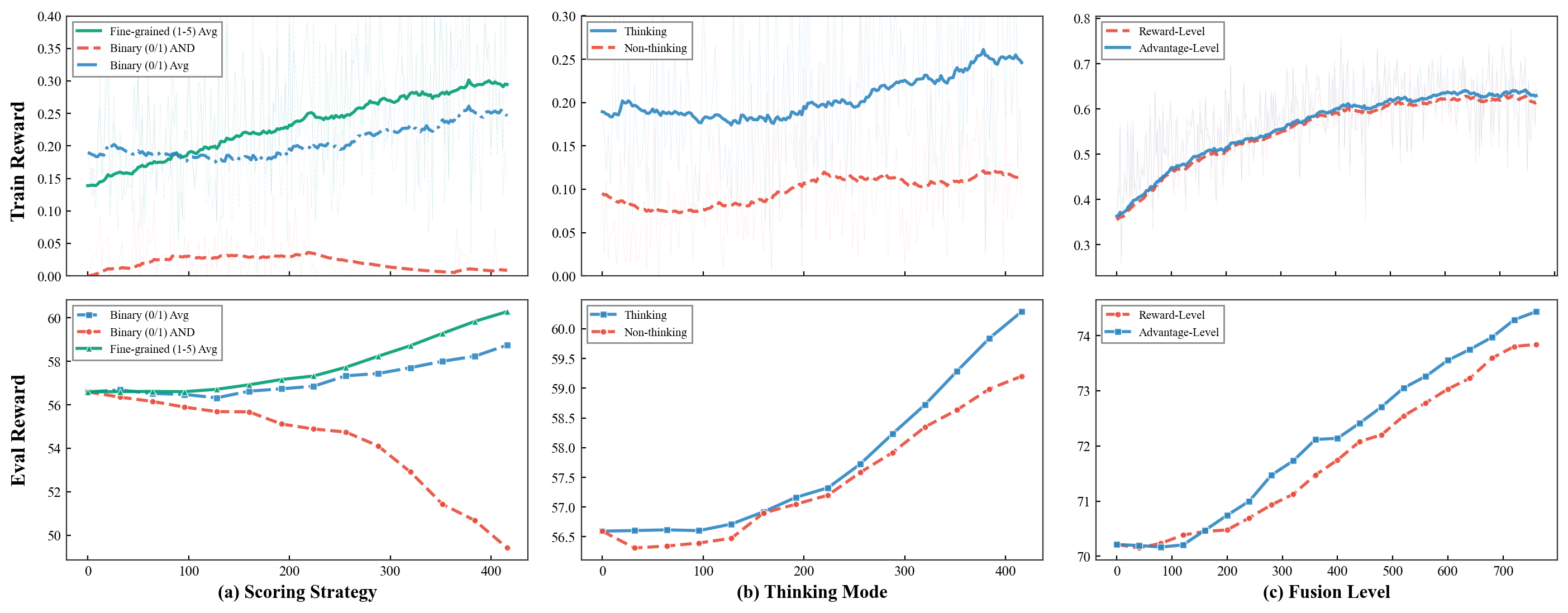}
    \caption{\textbf{Ablation on Reward Signal Design.} 
    The top row shows training reward, and the bottom row shows held-out evaluation reward. Fine-grained 1--5 scoring provides a more effectivenss learning signal than binary variants, thinking mode produces better performance with more accurate but higher-variance rewards, and advantage-level fusion consistently outperforms reward-level fusion.}
    \label{fig:reward_design}
\end{figure*}

\begin{figure*}[t]
    \centering
    \includegraphics[width=1\linewidth]{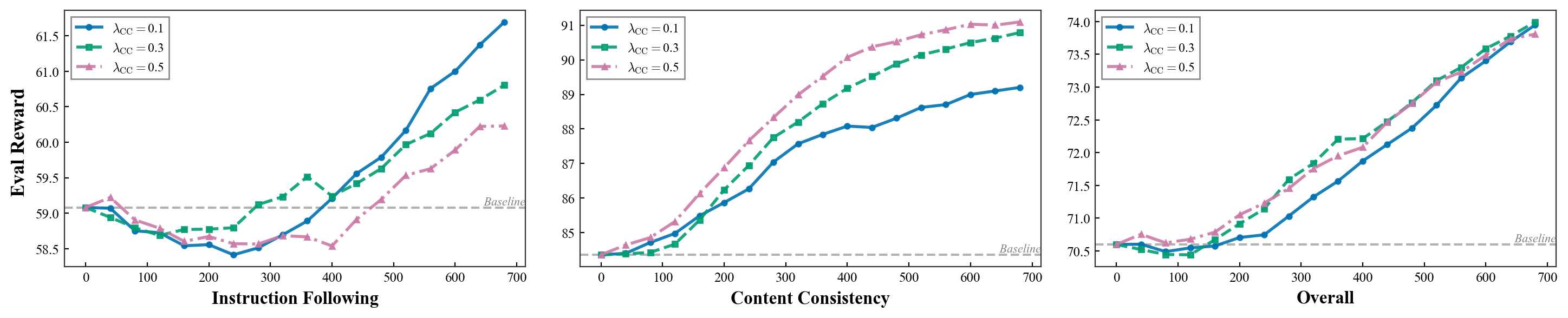}
    \caption{\textbf{Ablation on $\lambda_{\mathrm{CC}}$.} 
    Higher $\lambda_{\mathrm{CC}}$ improves CC at the cost of IF; $\lambda_{\mathrm{CC}}=0.3$ achieves the best overall balance.}
    \label{fig:flux2_nft_lambda_cc}
\end{figure*}

\begin{table}[t]
\centering
\caption{\textbf{Comparison of MT-EditFlow variants on FLUX.1-Kontext-dev.}}\label{tab:rl}
\resizebox{0.95\linewidth}{!}{
\begin{tabular}{l ccc ccc ccc}
\toprule
Method 
& \multicolumn{3}{c}{EdiVal-IF} 
& \multicolumn{3}{c}{EdiVal-CC} 
& \multicolumn{3}{c}{EdiVal-O} \\
\cmidrule(lr){2-4} \cmidrule(lr){5-7} \cmidrule(lr){8-10}
& T1 & T2 & T3 & T1 & T2 & T3 & T1 & T2 & T3 \\
\midrule
 MT-EditFlow-GRPO  
& 60.66 & 35.49 & 21.50 
& 95.35 & 92.26 & 90.24 
& 76.04 & 57.22 & 44.04 \\

\rowcolor{myblue}
MT-EditFlow-NFT 
& \textbf{62.24} & \textbf{38.99} & \textbf{22.90} 
& \textbf{95.75} & \textbf{92.47} & \textbf{90.62} 
& \textbf{77.20} & \textbf{60.04} & \textbf{45.56} \\

\bottomrule
\end{tabular}
}
\vspace{-0.3cm}
\end{table}

\subsection{Main Results}

Our experiments reveal several important insights regarding MT-EditFlow:

\begin{enumerate}
    \item \textbf{Substantial Multi-turn Improvement:} As shown in Tab.~\ref{tab:overall}, MT-EditFlow significantly improves overall multi-turn performance (EdiVal-O) over the corresponding open-source backbones, narrowing the gap to strong closed-source systems. The gains are particularly pronounced at later turns, with a 6.85-point increase for FLUX.1-Kontext-dev and a 2.90-point increase for FLUX.2-klein-base-9B.

    \item \textbf{Single-turn Performance:} MT-EditFlow also enhances single-turn editing performance, as shown in Tab.~\ref{tab:imgedit-bench}. These results demonstrate that our RL approach \textit{generalizes} well to different reference image sources and benchmarks.

    \item \textbf{Effect of Training Turns:} Fig.~\ref{fig:t3} further shows that the EdiVal-O (T3) score increases significantly even with only one training turn ($K=1$), indicating that single-turn training alone can be effective for multi-turn tasks. However, to achieve optimal multi-turn performance, additional RL training turns are required.

    \item \textbf{Reduced Exposure Bias:} Fig.~\ref{fig:task_rate} shows that MT-EditFlow maintains a much flatter marginal task success across turns compared to open-source baselines, approaching the trajectory-level reliability of strong closed-source models. This indicates that MT-EditFlow improves performance across the entire editing trajectory rather than merely boosting early-turn results.
\end{enumerate}

\subsection{Analysis on Reward Signal Design}
\label{sec:reward_model}
\label{sec:ablation_reward}
We study reward signal design along three axes: scoring granularity, evaluator prompting, and fusion level for joint IF+CC optimization.  

\subsubsection{Scoring strategy: granularity and sparsity}
We compare three turn-wise aggregation methods for $K=3$ turns: \textbf{Binary Avg} (per-turn 0/1 mean), \textbf{Binary AND} (1 only if all turns succeed), and \textbf{Fine-grained Avg} (per-turn 1--5 scores). Binary AND is extremely sparse: a trajectory is positive with probability $p^K$, yielding an ineffective learning signal. Fine-grained scoring provides partial credit, reduces sparsity, and stabilizes updates. Fig.~\ref{fig:reward_design}(a) shows it yields the strongest and most stable improvement; binary averaging is weaker, and AND degrades due to sparse rewards.

\subsubsection{Thinking vs.\ non-thinking mode}
We compare a \textbf{thinking} evaluator (structured per-turn analysis) with a \textbf{non-thinking} evaluator (direct score output). Fig.~\ref{fig:reward_design}(b) shows thinking mode improves the performance All main experiments use fine-grained 1--5 scoring with thinking prompting.

\subsubsection{Fusion level: reward vs.\ advantage}
Fig.~\ref{fig:reward_design}(c) shows advantage-level fusion outperforms reward-level fusion, with the gap widening during training.  Standardizing advantages prevents IF or CC from dominating due to scale differences.

\subsection{Ablation Studies}\label{sec:ablation}

\subsubsection{Ablations on different RL algorirthms} Tab.~\ref{tab:rl} shows that NFT outperforms GRPO on FLUX.1. In our implementation, Flow-GRPO fails on FLUX.2 because its likelihood estimation becomes inaccurate when the CFG guidance scale is embedded in the Diffusion Transformer. In contrast, NFT is more robust to different guidance designs with flow matching, making it the preferred choice.

\subsubsection{Impact of Content Consistency Weight $\lambda_{\mathrm{CC}}$}
Fig.~\ref{fig:flux2_nft_lambda_cc} visualizes this trade-off on FLUX.2 with DiffusionNFT: increasing $\lambda_{\mathrm{CC}}$ improves CC in both training and held-out evaluation, but suppresses IF, and the overall curve favors moderate weights. In sum, $\lambda_{\text{CC}}$ provides an effective way to prevent reward hacking. 




\begin{figure*}[t]
    \centering
    \includegraphics[width=1\linewidth]{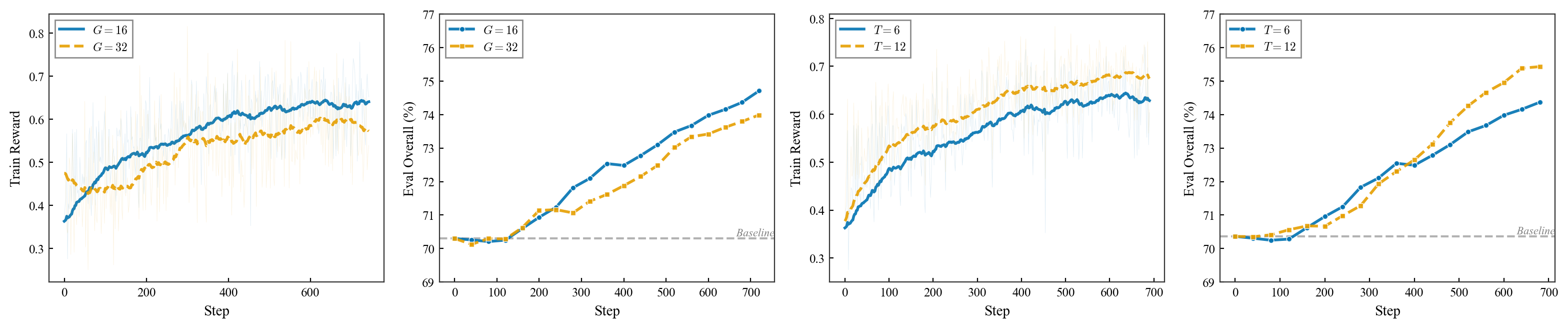}
    \caption{\textbf{Ablation on group size $G$ and discretization steps $T$.} We found that a smaller group size $G=16$ and a larger discretization step $T=12$ lead to better performance.
    }
    \label{fig:flux2_nft_num_steps}
    \vspace{-0.6cm}
\end{figure*}

\subsubsection{Group Size $G$ and Discretization Steps $T$}
For MT-EditFlow-NFT, Fig. \ref{fig:flux2_nft_num_steps} shows that the best performance is achieved with a moderate group size $G=16$ and finer discretization $T=12$.


\section{Limitations}
First, computational overhead is a concern, as using high-capacity VLMs for zero-shot rewards increases RL training latency compared to smaller reward models. 
Second, evaluator bias persists despite chain-of-thought prompting, as VLM hallucinations can introduce noise into the RL signal. 
Third, actor–critic RL methods could be explored for multi-turn editing, but they would likely incur additional computational cost due to the need for training a critic model, making them more expensive than trajectory-level advantage broadcasting. Finally, trajectory scaling remains an open question, as current evaluations focus on three-turn sequences and performance on longer interaction chains remains unexplored.

\section{Conclusion}
In this paper, we presented \textbf{MT-EditFlow}, a reinforcement learning framework specifically designed to address the challenges of multi-turn image editing within the flow-matching paradigm. By shifting from a single-turn, single-reward perspective to a multi-turn, multi-reward formulation, our approach effectively mitigates the all-or-nothing requirement and the exposure bias that typically causes sequential edits to fail. Through a systematic analysis of reward signal design,   MT-EditFlow, applicable to both GRPO and NFT-based methods, achieves  superior performance in multi-turn editing. By enabling robust, iterative refinement, MT-EditFlow provides a foundation for more reliable and natural human-AI collaboration in visual content creation.

\clearpage  


%
%
\bibliographystyle{splncs04}
\bibliography{egbib}

\clearpage

\appendix
\renewcommand*{\theHsection}{appendix.\Alph{section}}


\section{Overview}
 \begin{figure}[h]
    \centering
    \includegraphics[width=\linewidth]{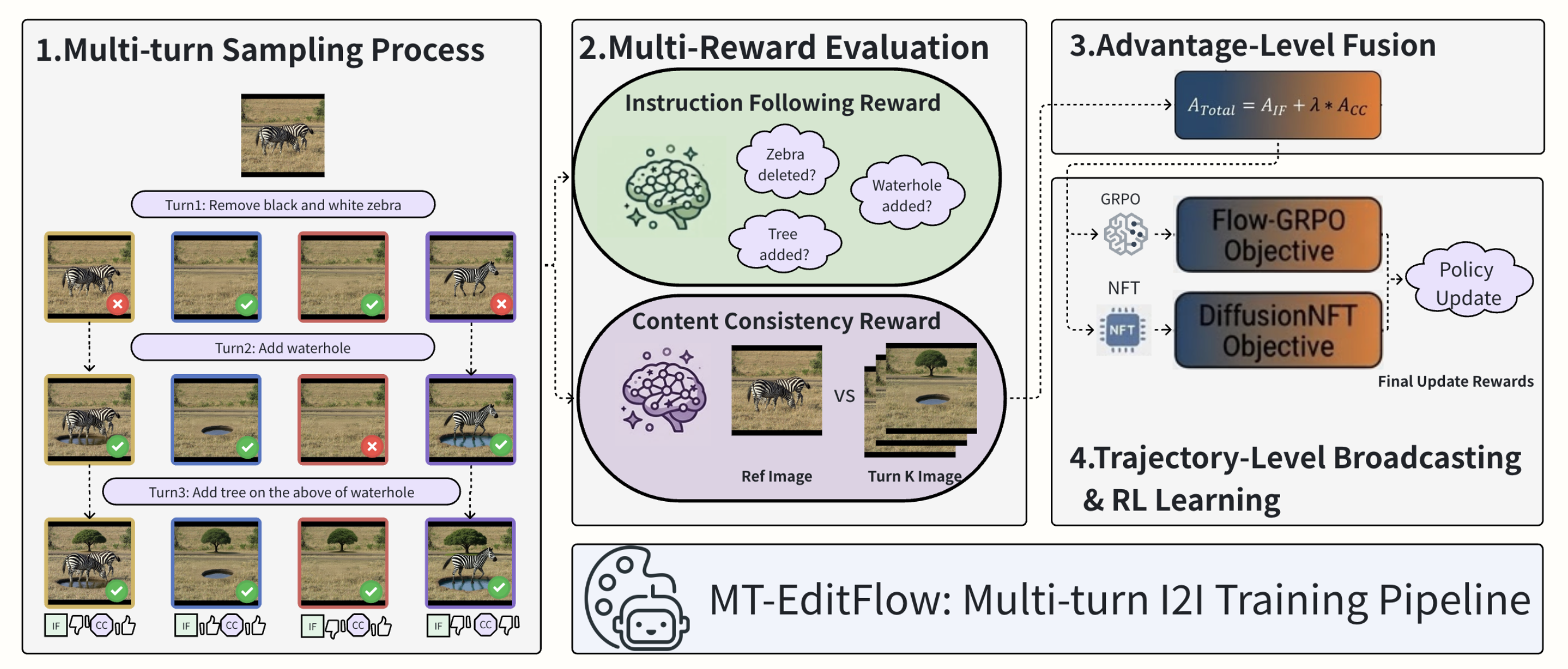}
    \caption{\textbf{Overview of MT-EditFlow.} The figure shows the overall MT-EditFlow pipeline, including multi-turn trajectory sampling and the corresponding training procedure.}
    \label{fig:appendix_main_overview}
\end{figure}
%

\section{Training Settings}

\subsection{Reward Model Setup}
We use Qwen3-VL-8B-Instruct as the automatic instruction-following reward model. During training, we query the multi-turn evaluator once per trajectory using the reference image, three turn edited images, and the concatenated three-turn instruction string, and average the returned turn-wise scores into a scalar IF reward. During evaluation, we instead score each turn independently with the single-turn evaluator. For content consistency, we use the same EdiVal CC pipeline described in the main paper and serve the CC models through a lightweight evaluation server during online RL, analogous to the IF reward service. The detailed CC computation is provided later in this supplementary. Unless otherwise stated, all main experiments use fine-grained 1--5 scoring with thinking mode.

\subsection{IF Prompt Format}
We provide the exact prompt/output template used in our experiments.

\subsubsection{Single-turn IF prompt.}
The single-turn IF prompt is mainly used for single-turn experiments. 
\begin{prompttemplate}{Single-turn IF prompt}
Original image:
[IMAGE]
Edited image:
[IMAGE]

Evaluate whether the edit instruction was properly executed by comparing the original and edited images.

Instruction: {}

A good edit should:
- Execute the requested changes accurately
- Preserve original content NOT mentioned in the instruction
- Avoid unintended changes beyond the instruction

Scoring (1-5):
1. Failed: Instruction completely ignored/opposite changes made, or critical original content destroyed
2. Minimal: Only minor parts of instruction followed, major elements missing/wrong, or severe content loss/unintended changes
3. Partial: Key instruction elements followed but incomplete/inaccurate, with noticeable original content loss/unintended modifications
4. Mostly Compliant: Instruction largely executed correctly with minor flaws, original content well-preserved with minimal unintended changes
5. Fully Compliant: Instruction accurately and completely executed, all non-targeted original content perfectly preserved

Provide analysis in <Thought> tag covering: instruction execution accuracy and completeness, preservation of non-targeted elements, unintended changes. Then give final score 1-5 in <Score>

<Thought>
[Analysis here]
</Thought>
<Score>X</Score>
\end{prompttemplate}

\subsubsection{Multi-turn IF prompt.}
\begin{prompttemplate}{Multi-turn IF prompt}
Original image:
[IMAGE]
Edited image:
[IMAGE][IMAGE][IMAGE]

Evaluate whether the edit instructions were properly executed by comparing the original and edited images across multiple turns.

Instructions: {}

There are 3 editing turns in total. Evaluate each turn separately based on:
- Execute the requested changes accurately
- Preserve original content NOT mentioned in the instruction
- Avoid unintended changes beyond the instruction

Scoring for each turn (1-5):
1. Failed: Instruction completely ignored/opposite changes made, or critical original content destroyed
2. Minimal: Only minor parts of instruction followed, major elements missing/wrong, or severe content loss/unintended changes
3. Partial: Key instruction elements followed but incomplete/inaccurate, with noticeable original content loss/unintended modifications
4. Mostly Compliant: Instruction largely executed correctly with minor flaws, original content well-preserved with minimal unintended changes
5. Fully Compliant: Instruction accurately and completely executed, all non-targeted original content perfectly preserved

Provide analysis in <Thought> tag covering each turn's execution. Then give scores in format:

<Thought>
[Turn 1 analysis]
[Turn 2 analysis]
[Turn 3 analysis]
</Thought>
<Turn1Score>X</Turn1Score>
<Turn2Score>X</Turn2Score>
<Turn3Score>X</Turn3Score>
<TotalScore>X</TotalScore>
\end{prompttemplate}

We compute the scalar IF reward by averaging the per-turn scores and normalizing to $[0,1]$.

\subsection{Computational Note}

Table~\ref{tab:appendix_runtime} summarizes representative wall-clock cost on NVIDIA B200 GPUs. Since our runs typically saturate by roughly 700 training steps, we report time-to-$\sim$700 effective steps rather than total run length. These numbers should be interpreted as indicative runtime measurements rather than strict efficiency benchmarks, since the runs differ in backbone size and RL objective.
\begin{table}[h]
\centering
\small
\caption{\textbf{Representative training cost to $\sim$700 effective steps on NVIDIA B200.}}
\label{tab:appendix_runtime}
\begin{tabular}{lccc}
\toprule
Run & Hours to $\sim$700 Steps & Effective Steps & Hours / 100 Steps \\
\midrule
MT-EditFlow (FLUX.1, GRPO) & 34.33 & 700 & 4.90 \\
MT-EditFlow (FLUX.1, NFT) & 21.37 & 684 & 3.12 \\
MT-EditFlow (FLUX.2, NFT) & 20.37 & 700 & 2.91 \\
\bottomrule
\end{tabular}
\end{table}
\section{Training Dataset Details}

\begin{figure}[t]
    \centering
    \begin{minipage}[t]{0.42\linewidth}
        \centering
        \includegraphics[width=\linewidth]{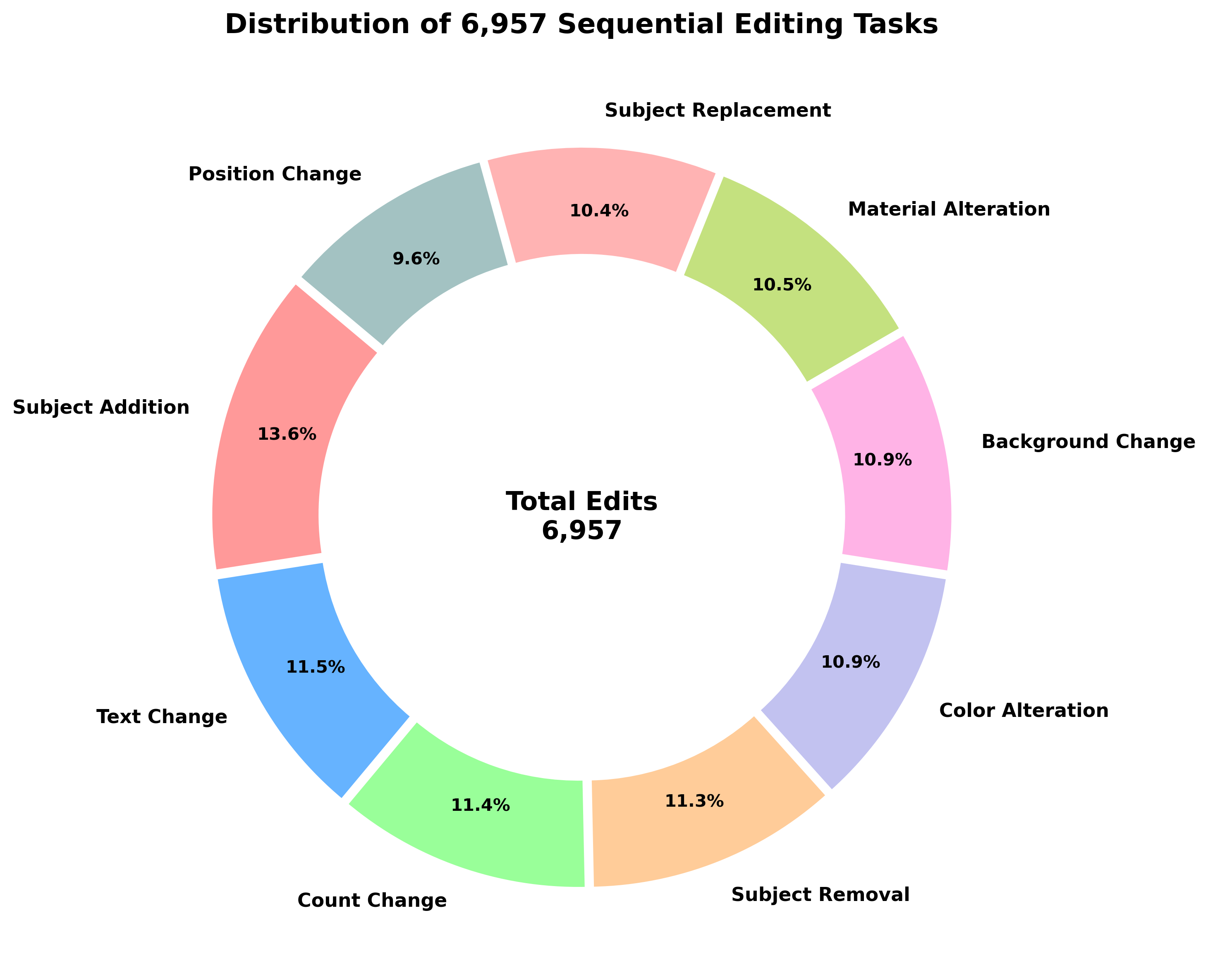}
    \end{minipage}
    \hfill
    \begin{minipage}[t]{0.54\linewidth}
        \centering
        \includegraphics[width=\linewidth]{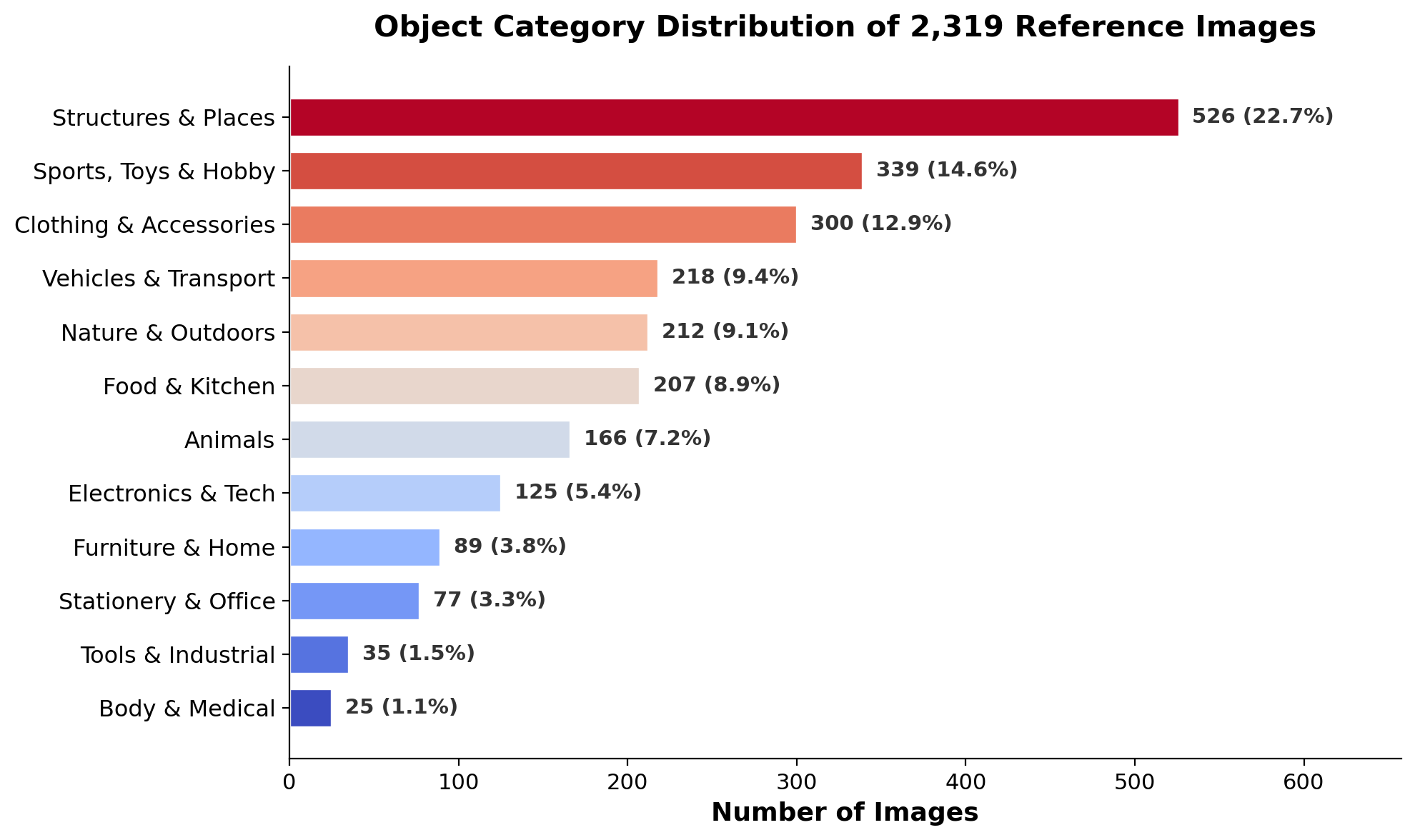}
    \end{minipage}
    \caption{\textbf{Training data distribution.} Left: distribution of 6,957 sequential edit steps across 9 editing categories. Right: object-category distribution of 2,319 reference images across 12 semantic categories.}
    \label{fig:task_dist_appendix}
\end{figure}

Our multi-turn training set contains \textbf{2,319 reference images} with planned \textbf{3-turn} editing trajectories, yielding \textbf{6,957 turn-level steps} for online RL fine-tuning. For analysis, we group the training trajectories into 9 major editing categories and track the semantic object categories appearing in the reference images. Fig.~\ref{fig:task_dist_appendix} summarizes these distributions: the left panel shows the 9 editing categories, and the right panel shows object-category coverage across the reference images.

\section{Algorithms}

Both MT-EditFlow variants share the same multi-turn rollout structure. For a prompt chain $\mathcal{C}=[c_1,\ldots,c_K]$, rollout $i$ sequentially generates edited images $\{x_{0,i}^{(k)}\}_{k=1}^{K}$, where each turn contains $T$ denoising steps and the rollout group size is $G$. After all $K$ turns are completed, we compute trajectory-level IF and CC rewards, convert them into the fused trajectory advantage $A_i^{\text{MT-EditFlow}}$, and broadcast the same signal to all turn-step positions $(k,t)$ inside the trajectory.
\begin{equation}
r_{i,\mathrm{IF}}
=
\frac{1}{K}\sum_{k=1}^{K}\Phi\!\left(x_{0,i}^{(k)},x_{0,i}^{(k-1)},c_k\right),
\qquad x_{0,i}^{(0)}:=x_f,
\label{eq:appendix_traj_if}
\end{equation}
\begin{equation}
\Omega_i^{(k)}=(\mathcal{U}_i^{(k)},\mathcal{O}_i^{(k)}),
\label{eq:appendix_cc_sets}
\end{equation}
\begin{equation}
I_i^{(0)} := x_f,
\qquad
I_i^{(k)} := x_{0,i}^{(k)},
\label{eq:appendix_cc_images}
\end{equation}
\begin{equation}
r_{i,\mathrm{CC}}
=
\frac{1}{K}\sum_{k=1}^{K}\mathrm{CC}\!\left(I_i^{(k)},I_i^{(k-1)},\Omega_i^{(k)}\right)
=
\frac{1}{2K}\sum_{k=1}^{K}\left(
s_{i,\mathrm{bg}}^{(k)}
+ \frac{1}{|\mathcal{U}_i^{(k)}|}\sum_{o\in\mathcal{U}_i^{(k)}} s_{i,o}^{(k)}
\right),
\label{eq:appendix_traj_cc}
\end{equation}
\begin{equation}
\Omega_{i,\mathrm{bg}}^{(k)}
=
\Omega - \bigcup_{o\in\mathcal{O}_i^{(k)}}\left(B_{i,o}^{(k-1)} \cup B_{i,o}^{(k)}\right),
\label{eq:appendix_cc_bg_region}
\end{equation}
\begin{align}
    \Omega_{i,\mathrm{o}}^{(k)}
= \bigcap_{o\in\mathcal{O}_i^{(k)}}\left(B_{i,o}^{(k-1)} \cup B_{i,o}^{(k)}\right),
\end{align}
\begin{equation}
s_{i,o}^{(k)}
=
\phi\!\left(I_{i,o}^{(k-1)}, I_{i,o}^{(k)}\right),
\qquad
I_{i,o}^{(k-1)} = \Omega_{i,\mathrm{o}}^{(k)} \odot I_i^{(k-1)},\;
I_{i,o}^{(k)} = \Omega_{i,\mathrm{o}}^{(k)} \odot I_i^{(k)},
\label{eq:appendix_cc_obj}
\end{equation}
\begin{equation}
s_{i,\mathrm{bg}}^{(k)}
=
\phi\!\left(I_{i,\mathrm{bg}}^{(k-1)}, I_{i,\mathrm{bg}}^{(k)}\right),
\qquad
I_{i,\mathrm{bg}}^{(k-1)} = \Omega_{i,\mathrm{bg}}^{(k)} \odot I_i^{(k-1)},\;
I_{i,\mathrm{bg}}^{(k)} = \Omega_{i,\mathrm{bg}}^{(k)} \odot I_i^{(k)},
\label{eq:appendix_cc_bg}
\end{equation}
where $\Phi$ is the IF evaluator, $\phi$ denotes the DINOv3-based similarity used by the CC metric, $\mathcal{U}_i^{(k)}$ is the set of unchanged objects at turn $k$, and $\mathcal{O}_i^{(k)}$ is the set of all objects considered by the CC metric at turn $k$. Unlike the simplified final-image view, the actual trajectory-level CC reward is computed turn by turn by comparing consecutive image pairs $(I_i^{(k-1)}, I_i^{(k)})$, i.e., reference vs.\ turn 1, turn 1 vs.\ turn 2, and turn 2 vs.\ turn 3. Here $B_{i,o}^{(k-1)}$ and $B_{i,o}^{(k)}$ denote the bounding boxes of object $o$ detected on the two consecutive images being compared. Eq.~\ref{eq:appendix_cc_bg_region} defines the background region by subtracting the union of detections from the two consecutive images from the full image area $\Omega$, Eq.~\ref{eq:appendix_cc_bg} measures background consistency on that region, and Eq.~\ref{eq:appendix_cc_obj} measures per-object consistency for unchanged objects. The final reported CC score follows the same code path as the evaluator and averages the DINOv3 object-consistency score (`object\_dinov3\_consistency\_mean`) with the masked DINOv3 background similarity (`background\_dinov3\_consistency`).

In the reward-level view, the resulting scalar trajectory reward is
\begin{equation}
r_i^{\text{MT-EditFlow}}
=
\lambda_{\mathrm{IF}}r_{i,\mathrm{IF}}+\lambda_{\mathrm{CC}}r_{i,\mathrm{CC}}.
\label{eq:appendix_traj_reward}
\end{equation}
For advantage-level fusion, we first compute group-relative advantages for each reward component,
\begin{equation}
    A_{i,m}
    =
    \frac{r_{i,m} - \frac{1}{G}\sum_{j=1}^{G} r_{j,m}}
    {\mathrm{std}(\{r_{j,m}\}_{j=1}^{G})},
    \qquad m \in \{\mathrm{IF}, \mathrm{CC}\},
    \label{eq:appendix_metric_adv}
\end{equation}
and then fuse them as
\begin{equation}
    A_i^{\text{MT-EditFlow}}
    =
    \lambda_{\mathrm{IF}}A_{i,\mathrm{IF}}+\lambda_{\mathrm{CC}}A_{i,\mathrm{CC}}.
    \label{eq:appendix_adv_fusion}
\end{equation}

\subsection{MT-EditFlow with Flow-GRPO}
MT-EditFlow extends standard Flow-GRPO by adding an outer turn loop while keeping the original clipped surrogate update. It augments the rollout structure with multi-turn trajectories and trajectory-level credit assignment.
{\small
\begin{equation}
\begin{aligned}
\mathcal{J}_{\text{MT-EditFlow-GRPO}}(\theta)
&=
\frac{1}{GKT}\sum_{i=1}^{G}\sum_{k=1}^{K}\sum_{t=1}^{T}
\Bigl[
\mathcal{L}_{\mathrm{clip}}\!\left(\rho_{t,i}^{(k)},A_i^{\text{MT-EditFlow}}\right) \\
&\qquad
-\beta_{\mathrm{KL}}\,\mathbb{D}_{\mathrm{KL}}(\pi_{\theta}\|\pi_{\mathrm{ref}})
\Bigr],
\end{aligned}
\label{eq:appendix_flow_grpo}
\end{equation}
}
with
\begin{equation}
\mathcal{L}_{\mathrm{clip}}(\rho,A)
=
\min\!\left(\rho A,\,
\mathrm{clip}(\rho,1-\epsilon,1+\epsilon)A\right).
\end{equation}
where the step-wise importance ratio is
\begin{equation}
\rho_{t,i}^{(k)}=
\frac{p_{\theta}(x_{t-\Delta t,i}^{(k)} \mid x_{t,i}^{(k)},x_f, c_k)}
{p_{\theta_{\mathrm{old}}}(x_{t-\Delta t,i}^{(k)} \mid x_{t,i}^{(k)},x_f, c_k)},
\label{eq:appendix_flow_ratio}
\end{equation}
In practice, $A_i^{\text{MT-EditFlow}}$ is obtained by computing group-relative advantages for $r_{i,\mathrm{IF}}$ and $r_{i,\mathrm{CC}}$ separately and then fusing them at the advantage level.

\refstepcounter{algorithmctr}\label{alg:mt_flow_grpo}
\begin{algobox}{Algorithm \thealgorithmctr: MT-EditFlow-GRPO}
\small
\noindent\parbox[t]{\linewidth}{\textbf{Require:} reference image $x_f$, prompt chain $\mathcal{C}=[c_1,\ldots,c_K]$, number of turns $K$, group size $G$, denoising steps $T$, rollout policy $\pi_{\theta_{\mathrm{old}}}$, training policy $\pi_\theta$.}\par
\noindent\parbox[t]{\linewidth}{\textbf{Initialize:} data buffer $\mathcal{D}\leftarrow \emptyset$.}\par
\noindent\makebox[2.2em][r]{1:}\hspace{0.35em}\textbf{for each iteration} $n$ \textbf{do}\par
\noindent\makebox[2.2em][r]{2:}\hspace{0.35em}\hspace*{1.2em}\parbox[t]{0.86\linewidth}{\textbf{for each sampled prompt chain} $\mathcal{C}$ \textbf{do} \hfill {\color{gray}\itshape // Rollout Step, Data Collection}}\par
\noindent\makebox[2.2em][r]{3:}\hspace{0.35em}\hspace*{2.4em}\textbf{for each rollout} $i=1,\ldots,G$ \textbf{do}\par
\noindent\makebox[2.2em][r]{4:}\hspace{0.35em}\hspace*{3.6em}Initialize the trajectory by $x_{0,i}^{(0)}\leftarrow x_f$.\par
\noindent\makebox[2.2em][r]{5:}\hspace{0.35em}\hspace*{3.6em}\textbf{for} $k=1,\ldots,K$ \textbf{do}\par
\noindent\makebox[2.2em][r]{6:}\hspace{0.35em}\hspace*{4.8em}\parbox[t]{0.76\linewidth}{Sample a $T$-step stochastic denoising trajectory from $\pi_{\theta_{\mathrm{old}}}$ conditioned on $(x_f,c_k)$.}\par
\noindent\makebox[2.2em][r]{7:}\hspace{0.35em}\hspace*{4.8em}Decode the turn output $x_{0,i}^{(k)}$ and feed it to the next turn.\par
\noindent\makebox[2.2em][r]{8:}\hspace{0.35em}\hspace*{3.6em}\textbf{end for}\par
\noindent\makebox[2.2em][r]{9:}\hspace{0.35em}\hspace*{3.6em}\parbox[t]{0.78\linewidth}{Compute $r_{i,\mathrm{IF}}$ by Eq.~\ref{eq:appendix_traj_if}, compute $r_{i,\mathrm{CC}}$ by Eqs.~\ref{eq:appendix_traj_cc}--\ref{eq:appendix_cc_bg}, and form $r_i^{\text{MT-EditFlow}}$ by Eq.~\ref{eq:appendix_traj_reward}.}\par
\noindent\makebox[2.2em][r]{10:}\hspace{0.35em}\hspace*{2.4em}\textbf{end for}\par
\noindent\makebox[2.2em][r]{11:}\hspace{0.35em}\hspace*{2.4em}\parbox[t]{0.82\linewidth}{Compute $A_{i,m}$ by Eq.~\ref{eq:appendix_metric_adv} and fuse $A_i^{\text{MT-EditFlow}}$ by Eq.~\ref{eq:appendix_adv_fusion}; store the resulting trajectory states in $\mathcal{D}$.}\par
\noindent\makebox[2.2em][r]{12:}\hspace{0.35em}\hspace*{1.2em}\textbf{end for}\par
\noindent\makebox[2.2em][r]{13:}\hspace{0.35em}\hspace*{1.2em}\parbox[t]{0.86\linewidth}{\textbf{for each mini-batch} $\mathcal{B}\subset\mathcal{D}$ \textbf{do} \hfill {\color{gray}\itshape // Gradient Step, Policy Optimization}}\par
\noindent\makebox[2.2em][r]{14:}\hspace{0.35em}\hspace*{2.4em}\parbox[t]{0.82\linewidth}{Broadcast $A_i^{\text{MT-EditFlow}}$ to the entire trajectory, obtain $\mathcal{J}_{\text{MT-EditFlow-GRPO}}(\theta)$ by Eq.~\ref{eq:appendix_flow_grpo}, and update $\theta$ using Adam.}\par
\noindent\makebox[2.2em][r]{15:}\hspace{0.35em}\hspace*{1.2em}\textbf{end for}\par
\noindent\makebox[2.2em][r]{16:}\hspace{0.35em}\hspace*{1.2em}\parbox[t]{0.84\linewidth}{Refresh the rollout policy periodically by $\theta_{\mathrm{old}}\leftarrow\theta$, and clear $\mathcal{D}\leftarrow\emptyset$. \hfill {\color{gray}\itshape // Online Update}}\par
\noindent\makebox[2.2em][r]{17:}\hspace{0.35em}\textbf{end for}\par
\noindent\textbf{Output:} $\pi_\theta$\par
\end{algobox}

\subsection{MT-EditFlow with DiffusionNFT}
MT-EditFlow extends DiffusionNFT in the same multi-turn manner. The novelty is again the trajectory construction and multi-turn reward aggregation, while the per-step NFT update remains a positive/negative implicit velocity-matching objective. For readability, we suppress the shared arguments $(x_{t,i}^{(k)}, t, x_f, c_k)$ of $v_\theta^{+}$ and $v_\theta^{-}$ below.

\begin{equation}
\mathcal{L}_{\mathrm{clip}}^{\mathrm{NFT}}(A)
=
\frac{1}{2} + \frac{1}{2}\,\mathrm{clip}\!\left(\frac{A}{z_c},-1,1\right),
\label{eq:appendix_nft_clip}
\end{equation}
\begin{equation}
v_{\theta}^{+}=(1-\beta)v_{\theta_{\mathrm{old}}}+\beta v_{\theta},
\qquad
v_{\theta}^{-}=(1+\beta)v_{\theta_{\mathrm{old}}}-\beta v_{\theta},
\label{eq:appendix_nft_pos_neg}
\end{equation}
\begin{equation}
\begin{aligned}
\mathcal{L}_{\text{MT-EditFlow-NFT}}
&=
\frac{1}{GKT}\sum_{i=1}^{G}\sum_{k=1}^{K}\sum_{t=1}^{T}
\Bigl[
\mathcal{L}_{\mathrm{clip}}^{\mathrm{NFT}}\!\left(A_i^{\text{MT-EditFlow}}\right) \|v_{\theta}^{+}-v\|_2^2 \\
&\qquad
+ \left(1-\mathcal{L}_{\mathrm{clip}}^{\mathrm{NFT}}\!\left(A_i^{\text{MT-EditFlow}}\right)\right)\|v_{\theta}^{-}-v\|_2^2
\Bigr],
\end{aligned}
\label{eq:appendix_nft}
\end{equation}
where $A_i^{\text{MT-EditFlow}}$ is the fused trajectory advantage from Eq.~\ref{eq:appendix_adv_fusion}. The same trajectory-level advantage is broadcast to all denoising steps and turns of rollout $i$.

\refstepcounter{algorithmctr}\label{alg:mt_nft}
\begin{algobox}{Algorithm \thealgorithmctr: MT-EditFlow-NFT}
\small
\noindent\parbox[t]{\linewidth}{\textbf{Require:} reference image $x_f$, prompt chain $\mathcal{C}=[c_1,\ldots,c_K]$, number of turns $K$, group size $G$, denoising steps $T$, rollout model $v_{\theta_{\mathrm{old}}}$, training model $v_\theta$.}\par
\noindent\parbox[t]{\linewidth}{\textbf{Initialize:} data buffer $\mathcal{D}\leftarrow \emptyset$.}\par
\noindent\makebox[2.2em][r]{1:}\hspace{0.35em}\textbf{for each iteration} $n$ \textbf{do}\par
\noindent\makebox[2.2em][r]{2:}\hspace{0.35em}\hspace*{1.2em}\parbox[t]{0.86\linewidth}{\textbf{for each sampled prompt chain} $\mathcal{C}$ \textbf{do} \hfill {\color{gray}\itshape // Rollout Step, Data Collection}}\par
\noindent\makebox[2.2em][r]{3:}\hspace{0.35em}\hspace*{2.4em}\textbf{for each rollout} $i=1,\ldots,G$ \textbf{do}\par
\noindent\makebox[2.2em][r]{4:}\hspace{0.35em}\hspace*{3.6em}Initialize the trajectory by $x_{0,i}^{(0)}\leftarrow x_f$.\par
\noindent\makebox[2.2em][r]{5:}\hspace{0.35em}\hspace*{3.6em}\textbf{for} $k=1,\ldots,K$ \textbf{do}\par
\noindent\makebox[2.2em][r]{6:}\hspace{0.35em}\hspace*{4.8em}\parbox[t]{0.76\linewidth}{Run a $T$-step forward/noise-conditioned rollout with $v_{\theta_{\mathrm{old}}}$ conditioned on $(x_f,c_k)$.}\par
\noindent\makebox[2.2em][r]{7:}\hspace{0.35em}\hspace*{4.8em}Decode the turn output $x_{0,i}^{(k)}$ and feed it to the next turn.\par
\noindent\makebox[2.2em][r]{8:}\hspace{0.35em}\hspace*{3.6em}\textbf{end for}\par
\noindent\makebox[2.2em][r]{9:}\hspace{0.35em}\hspace*{3.6em}\parbox[t]{0.78\linewidth}{Compute $r_{i,\mathrm{IF}}$ by Eq.~\ref{eq:appendix_traj_if}, compute $r_{i,\mathrm{CC}}$ by Eqs.~\ref{eq:appendix_traj_cc}--\ref{eq:appendix_cc_bg}, and form $r_i^{\text{MT-EditFlow}}$ by Eq.~\ref{eq:appendix_traj_reward}.}\par
\noindent\makebox[2.2em][r]{10:}\hspace{0.35em}\hspace*{2.4em}\textbf{end for}\par
\noindent\makebox[2.2em][r]{11:}\hspace{0.35em}\hspace*{2.4em}\parbox[t]{0.82\linewidth}{Compute $A_{i,m}$ by Eq.~\ref{eq:appendix_metric_adv} and fuse $A_i^{\text{MT-EditFlow}}$ by Eq.~\ref{eq:appendix_adv_fusion}; store the resulting trajectory states in $\mathcal{D}$.}\par
\noindent\makebox[2.2em][r]{12:}\hspace{0.35em}\hspace*{1.2em}\textbf{end for}\par
\noindent\makebox[2.2em][r]{13:}\hspace{0.35em}\hspace*{1.2em}\parbox[t]{0.86\linewidth}{\textbf{for each mini-batch} $\mathcal{B}\subset\mathcal{D}$ \textbf{do} \hfill {\color{gray}\itshape // Gradient Step, Policy Optimization}}\par
\noindent\makebox[2.2em][r]{14:}\hspace{0.35em}\hspace*{2.4em}\parbox[t]{0.82\linewidth}{Broadcast $A_i^{\text{MT-EditFlow}}$ to the entire trajectory, obtain $\mathcal{L}_{\text{MT-EditFlow-NFT}}$ by Eq.~\ref{eq:appendix_nft}, and update $\theta$ using Adam.}\par
\noindent\makebox[2.2em][r]{15:}\hspace{0.35em}\hspace*{1.2em}\textbf{end for}\par
\noindent\makebox[2.2em][r]{16:}\hspace{0.35em}\hspace*{1.2em}\parbox[t]{0.84\linewidth}{Refresh the rollout policy periodically by $\theta_{\mathrm{old}}\leftarrow\theta$, and clear $\mathcal{D}\leftarrow\emptyset$. \hfill {\color{gray}\itshape // Online Update}}\par
\noindent\makebox[2.2em][r]{17:}\hspace{0.35em}\textbf{end for}\par
\noindent\textbf{Output:} $v_\theta$\par
\end{algobox}

\section{Additional Results and Ablations}

\subsection{Why Fuse at the Advantage Level}
\begin{figure}[t]
    \centering
    \includegraphics[width=0.72\linewidth]{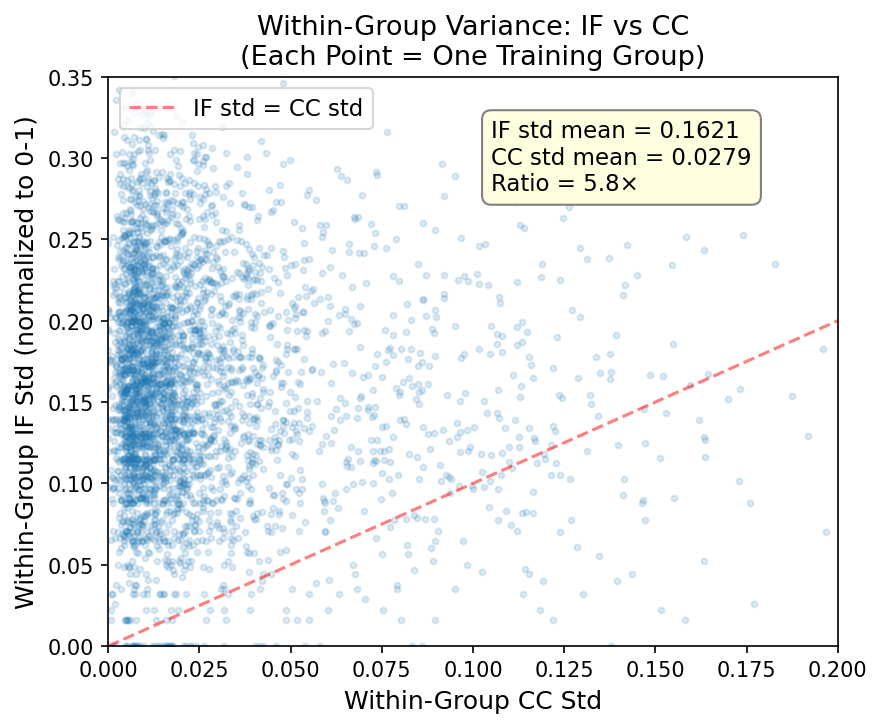}
    \caption{\textbf{Within-group reward variance analysis.} Raw IF and CC rewards can have substantially different within-group scales and variances. This mismatch is one of the main motivations for performing fusion at the advantage level rather than directly in raw reward space.}
    \label{fig:within_group_variance}
\end{figure}

Fig.~\ref{fig:within_group_variance} provides an empirical view of the motivation behind advantage-level fusion. Within the same rollout group, the raw IF and CC rewards can exhibit noticeably different scales and variances, so directly fusing them at the reward level makes the larger-variance component dominate the update. In our case, the within-group variance of IF is about 5.8$\times$ that of CC, which means reward-level fusion becomes effectively IF-dominated, making the fused update noisier and weakening the practical influence of CC despite its explicit weight. This observation motivates our design choice in the main paper: we first convert each reward into a group-relative advantage and then fuse them in the advantage space, where the two signals are more comparable and the resulting policy update is more stable.

\subsection{$\lambda_{\mathrm{CC}}$ Sweep on FLUX.1-Kontext-dev}
Table~\ref{tab:appendix_lambda_cc} complements the main-paper FLUX.2 sweep by showing the same trend on MT-EditFlow with FLUX.1: moderate $\lambda_{\mathrm{CC}}$ values provide the best overall balance, while overly large $\lambda_{\mathrm{CC}}$ pushes the model toward a minimal-edit consistency-hacking regime. In particular, when $\lambda_{\mathrm{CC}}=1.0$, the CC score is even slightly higher than the baseline (91.72 vs.\ 90.85), but IF drops sharply (50.44 vs.\ 56.07). This indicates a degenerate optimization path: compared with making large, correct edits that improve IF, it is much easier to increase CC by making only tiny changes, or even effectively doing nothing, so that most of the original image remains unchanged.

\begin{table}[h]
\centering
\small
\caption{\textbf{Impact of $\lambda_{\mathrm{CC}}$.} Best checkpoint (highest Overall) for each $\lambda_{\mathrm{CC}}$ on FLUX.1-Kontext-dev with Flow-GRPO. Overall $=\sqrt{\text{IF}\times\text{CC}}$.}
\label{tab:appendix_lambda_cc}
\begin{tabular}{ccccc}
\toprule
$\lambda_{\mathrm{CC}}$ & Best Step & IF (\%) & CC (\%) & Overall (\%) \\
\midrule
Baseline & -- & 56.07 & 90.85 & 71.37 \\
\midrule
0.1  & 352 & 61.72 & 90.09 & 74.57 \\
0.15 & 608 & 67.60 & 85.59 & \textbf{76.07} \\
0.2  & 512 & 64.22 & 88.66 & 75.46 \\
0.5  & 512 & 61.85 & 89.61 & 74.45 \\
1.0  & 224 & 50.44 & 91.72 & 68.02 \\
\bottomrule
\end{tabular}
\end{table}

\subsection{Group Size $G$ on FLUX.1-Kontext-dev}
\begin{figure}[t]
    \centering
    \includegraphics[width=0.92\linewidth]{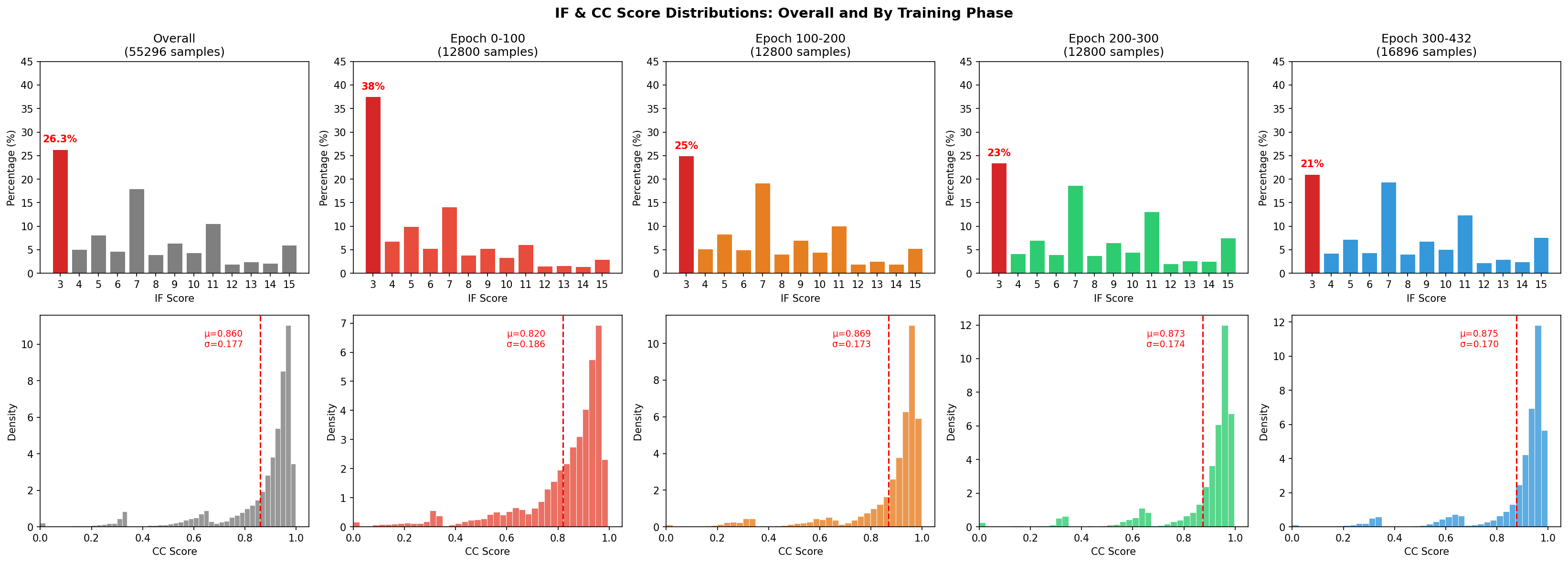}
    \caption{\textbf{IF/CC reward statistics related to group-size selection.} The IF signal occupies a small number of discrete score levels and dominates the fused advantage variance, while CC is comparatively smoother and lower-impact. This helps explain why a moderate group size can already provide a reliable within-group ranking signal, making the benefit of increasing $G$ beyond that point limited rather than guaranteed.}
    \label{fig:if_cc_distribution_appendix}
\end{figure}

Table~\ref{tab:appendix_group_size} shows a non-monotonic group-size trend on MT-EditFlow with FLUX.1. The main conclusion is not that larger $G$ is always worse, but that choosing an appropriate $G$ matters: a moderate group size already provides a strong ranking signal, and further increasing $G$ does not guarantee additional gains. Fig.~\ref{fig:if_cc_distribution_appendix} helps explain this behavior with concrete statistics. Under our advantage-level fusion,
\[
A_i^{\text{MT-EditFlow}} = 1.0 \cdot A_{i,\mathrm{IF}} + 0.3 \cdot A_{i,\mathrm{CC}},
\]
the weighted fused signal is strongly IF-dominated. In the analyzed run, the raw within-group variance of IF is about $3.1\times$ that of CC; after fusion with weights $(1.0, 0.3)$, the influence of CC is further downweighted in the update signal. At the same time, the IF reward is discrete and highly concentrated. Although its theoretical range is $[3,15]$, the empirical distribution is strongly bimodal: about $26.3\%$ of samples are at IF$=3$, while most of the remaining mass lies between 9 and 15, so the effective ranking signal is carried by only about 7--8 commonly occupied score levels. This means a moderate group size such as $G{=}16$ is often already sufficient to cover the useful within-group ordering. In GRPO, $G$ is mainly used to form a reliable within-group ranking, not to directly increase the update magnitude. Once this ranking is already stable, increasing $G$ mostly improves statistical stability, while under a fixed batch budget it also reduces the number of distinct prompts per update. Together, these observations suggest that an intermediate $G$ is a better practical operating point than simply increasing group size by default.

\begin{table}[t]
\centering
\small
\caption{\textbf{Ablation of group size $G$.} Best result (highest Overall) for each $G$ with $\lambda_{\mathrm{CC}}=0.2$ on FLUX.1-Kontext-dev with Flow-GRPO.}
\label{tab:appendix_group_size}
\begin{tabular}{cccc}
\toprule
$G$ & IF (\%) & CC (\%) & Overall (\%) \\
\midrule
8  & 64.64 & 87.72 & 75.30 \\
16 & 68.54 & 87.81 & \textbf{77.58} \\
32 & 64.22 & 88.66 & 75.46 \\
64 & 67.66 & 82.14 & 74.55 \\
\bottomrule
\end{tabular}
\end{table}

\subsection{Cross-turn Robustness}
Table~\ref{tab:appendix_cross_turn} reports the marginal task-success rate across turns and can be viewed as the numeric version of Fig.~\ref{fig:task_rate} in the main paper. All MT-EditFlow variants exhibit substantially smaller degradation than the corresponding open-source baselines, supporting the claim that trajectory-level training reduces exposure bias in multi-turn editing.

\begin{table}[t]
\centering
\small
\caption{\textbf{Cross-turn robustness.} Marginal task success rate drop from T1 to T3. MT-EditFlow shows a smaller cross-turn drop than open-source baselines.}
\label{tab:appendix_cross_turn}
\begin{tabular}{@{}lcccc@{}}
\toprule
Model & T1 & T2 & T3 & $\Delta$ \\
\midrule
GPT-Image-1 & 73.12 & 74.44 & 73.12 & +0.00 \\
FLUX.1-Kontext-max & 69.49 & 69.11 & 70.43 & +0.94 \\
FLUX.1-Kontext-dev & 59.97 & 56.29 & 51.40 & $-$8.57 \\
FLUX.2-klein-base-9B & 69.41 & 67.13 & 67.13 & $-$2.28 \\
\midrule
MT-EditFlow (FLUX.1, GRPO) & 63.99 & 62.76 & 62.41 & $-$1.58 \\
MT-EditFlow (FLUX.1, NFT) & 62.24 & 61.89 & 61.19 & $-$1.05 \\
MT-EditFlow (FLUX.2, NFT) & 70.28 & 69.23 & 69.93 & $-$0.35 \\
\bottomrule
\end{tabular}
\end{table}

\subsection{Visual Quality and Single-turn Generalization}
\begin{figure}[H]
    \centering
    \includegraphics[width=0.82\linewidth]{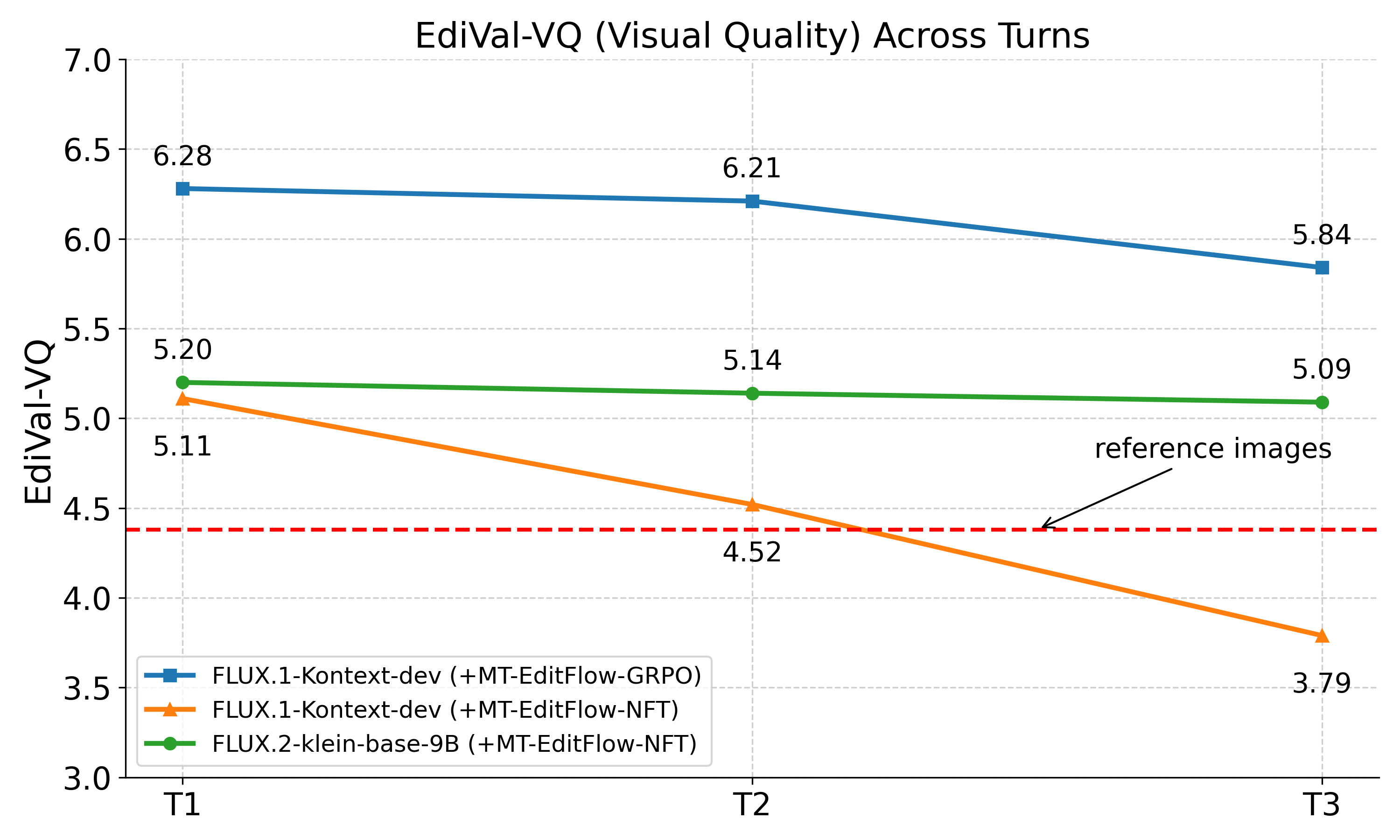}
    \caption{\textbf{Visual quality across turns.} Turn-level visual quality remains strong across the editing trajectory and generally stays above the reference-image baseline, indicating that MT-EditFlow improves multi-turn performance without obvious visual-quality collapse.}
    \label{fig:vq_appendix}
\end{figure}

Table~\ref{tab:overall} in the main paper shows that MT-EditFlow substantially improves multi-turn overall performance over the corresponding open-source backbones, with larger gains at later turns. These gains do not come from collapsing visual fidelity. Fig.~\ref{fig:vq_appendix} shows that turn-level visual quality remains above the reference-image baseline for almost all edits, indicating that stronger instruction following and consistency are not achieved by sacrificing image quality.

\section{Additional Qualitative Examples}

\begin{figure*}
    \centering
    \begin{tabular}{cccc}
        \includegraphics[width=0.22\linewidth]{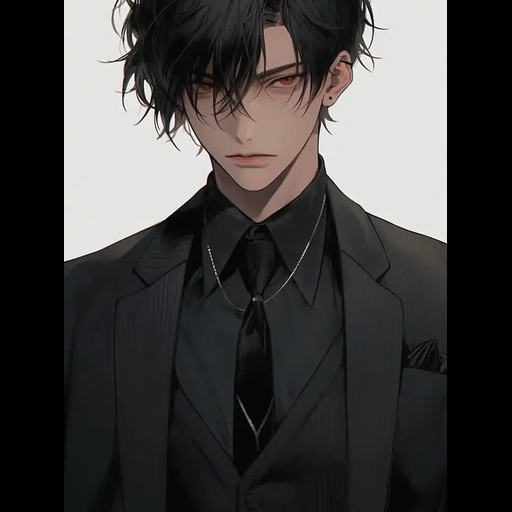} &
        \includegraphics[width=0.22\linewidth]{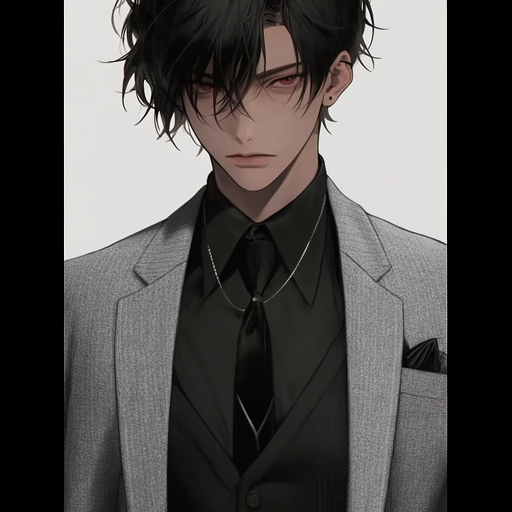} &
        \includegraphics[width=0.22\linewidth]{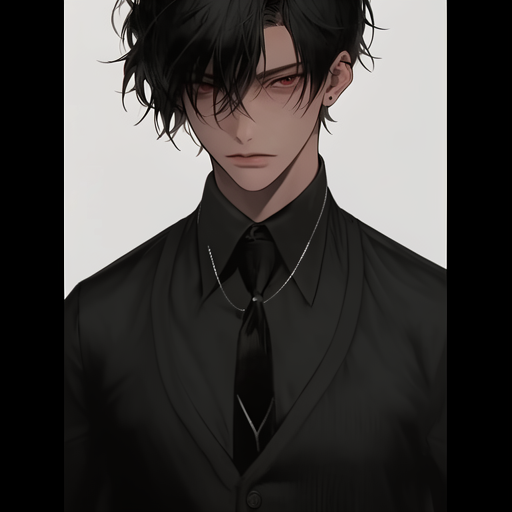} &
        \includegraphics[width=0.22\linewidth]{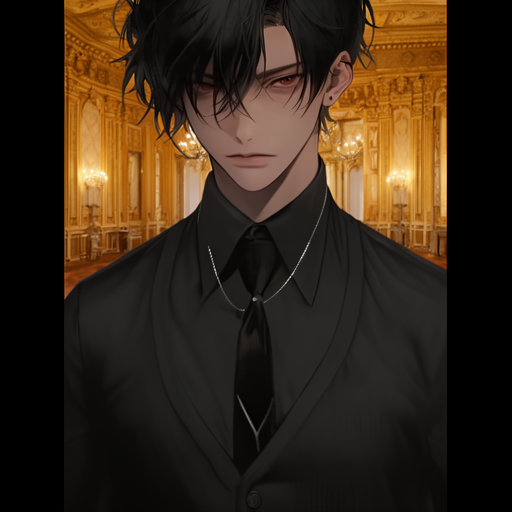} \\
        Input & Turn 1 & Turn 2 & Turn 3
    \end{tabular}

    \vspace{0.35em}
    \parbox{0.95\linewidth}{\footnotesize
    \textbf{Turn 1:} Change the color of black suit jacket to gray.\\
    \textbf{Turn 2:} Remove gray suit jacket.\\
    \textbf{Turn 3:} Change the background to ballroom, while keeping the gray shirt, black tie, and metal chain necklace unchanged.}

    \vspace{0.7em}
    \begin{tabular}{cccc}
        \includegraphics[width=0.22\linewidth]{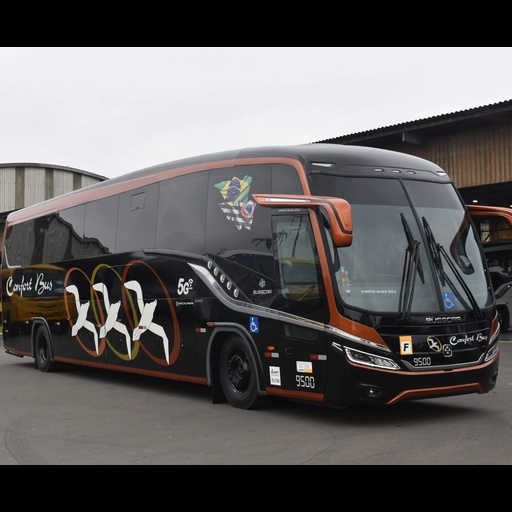} &
        \includegraphics[width=0.22\linewidth]{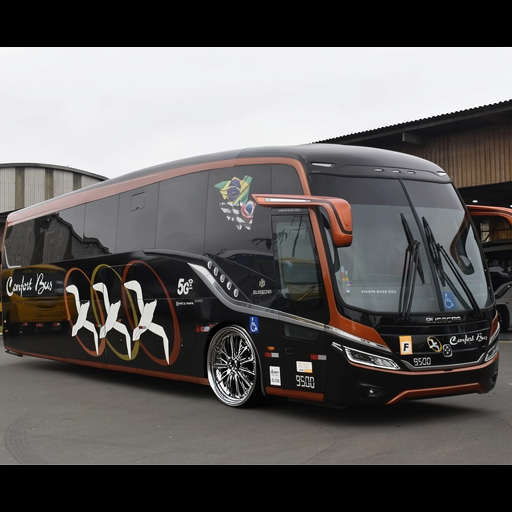} &
        \includegraphics[width=0.22\linewidth]{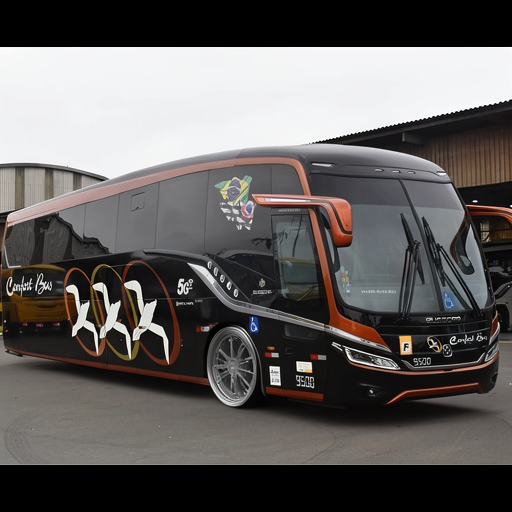} &
        \includegraphics[width=0.22\linewidth]{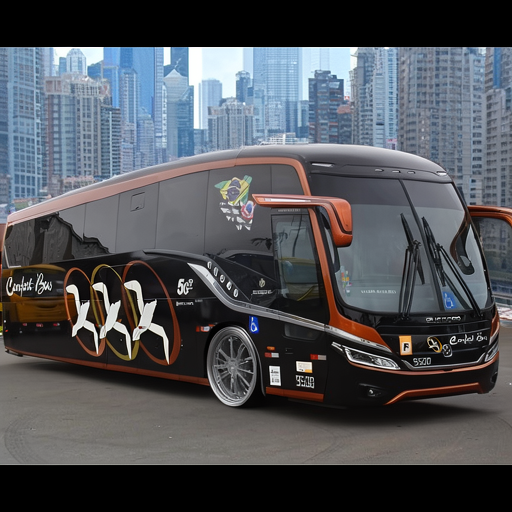} \\
        Input & Turn 1 & Turn 2 & Turn 3
    \end{tabular}

    \vspace{0.35em}
    \parbox{0.95\linewidth}{\footnotesize
    \textbf{Turn 1:} Replace rubber black tire with alloy wheel.\\
    \textbf{Turn 2:} Change the color of alloy wheel to gray.\\
    \textbf{Turn 3:} Change the background to cityscape, while keeping the gray alloy wheel unchanged.}

    \begin{tabular}{cccc}
        \includegraphics[width=0.22\linewidth]{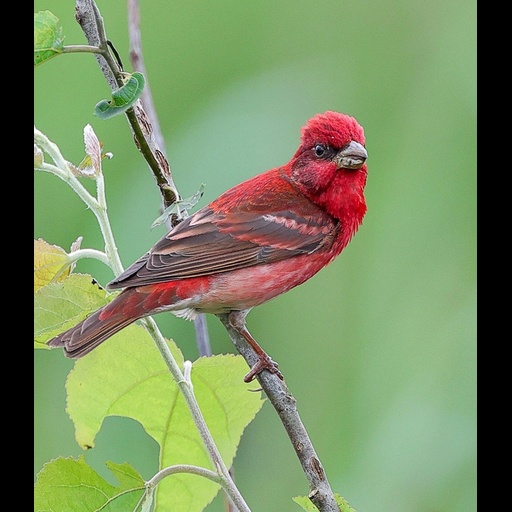} &
        \includegraphics[width=0.22\linewidth]{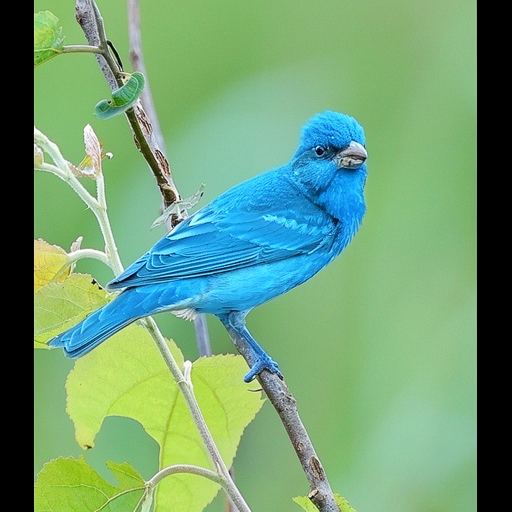} &
        \includegraphics[width=0.22\linewidth]{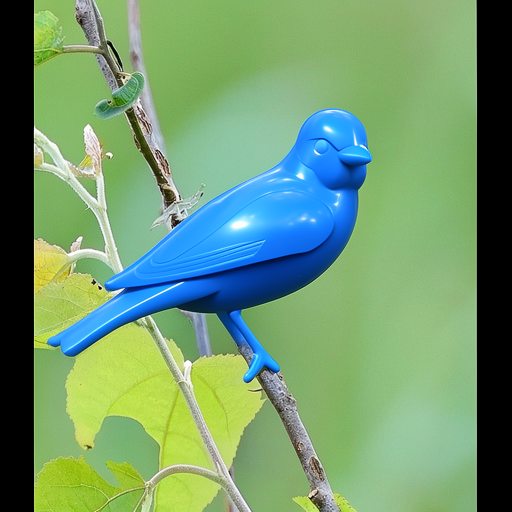} &
        \includegraphics[width=0.22\linewidth]{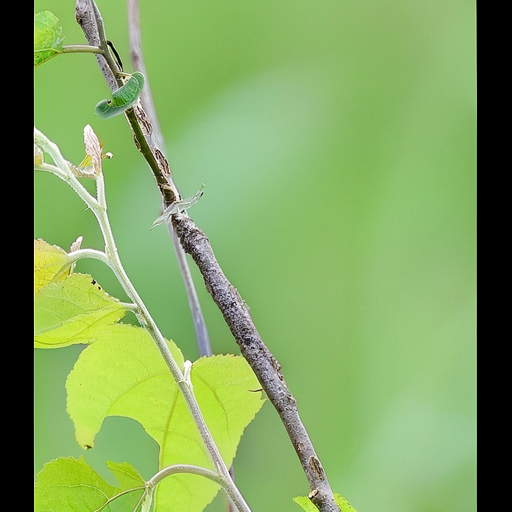} \\
        Input & Turn 1 & Turn 2 & Turn 3
    \end{tabular}

    \vspace{0.35em}
    \parbox{0.95\linewidth}{\footnotesize
    \textbf{Turn 1:} Change the color of feathered red bird to blue.\\
    \textbf{Turn 2:} Change the material of blue red bird to plastic.\\
    \textbf{Turn 3:} Remove plastic blue red bird.}

\end{figure*}

\vspace{0.8em}

\begin{figure*}
    \centering

    \vspace{0.7em}
    \begin{tabular}{cccc}
        \includegraphics[width=0.22\linewidth]{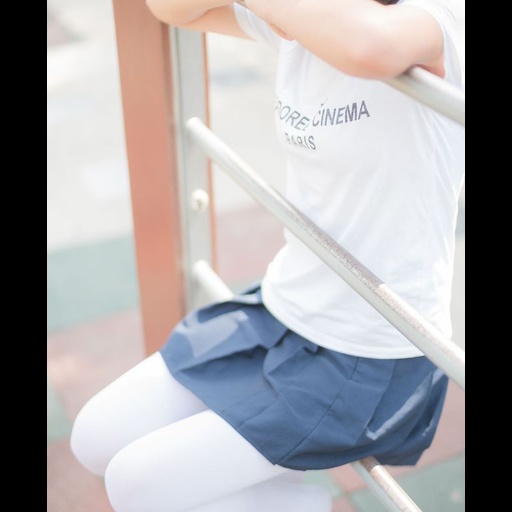} &
        \includegraphics[width=0.22\linewidth]{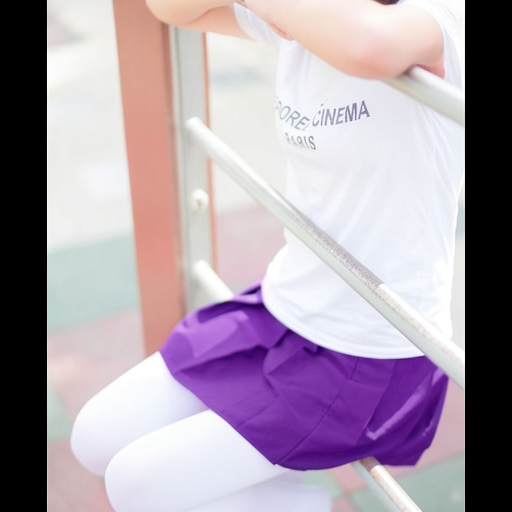} &
        \includegraphics[width=0.22\linewidth]{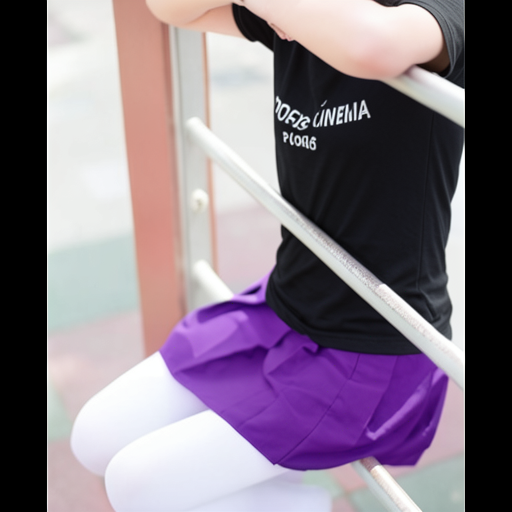} &
        \includegraphics[width=0.22\linewidth]{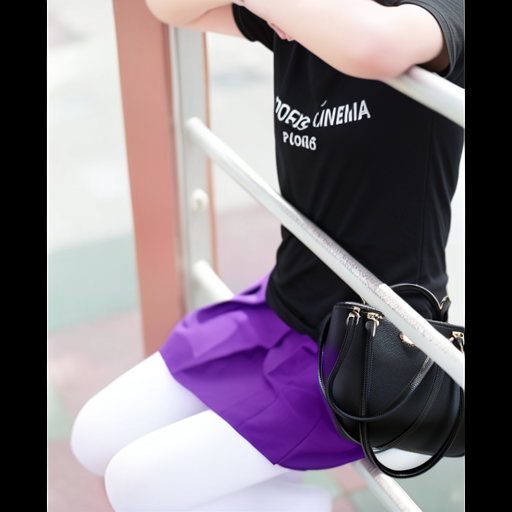} \\
        Input & Turn 1 & Turn 2 & Turn 3
    \end{tabular}

    \vspace{0.35em}
    \parbox{0.95\linewidth}{\footnotesize
    \textbf{Turn 1:} Change the color of cotton blue skirt to purple.\\
    \textbf{Turn 2:} Change the material of cotton white t-shirt to polyester.\\
    \textbf{Turn 3:} Add a bag to the right of the cotton purple skirt.}

    \begin{tabular}{cccc}
        \includegraphics[width=0.22\linewidth]{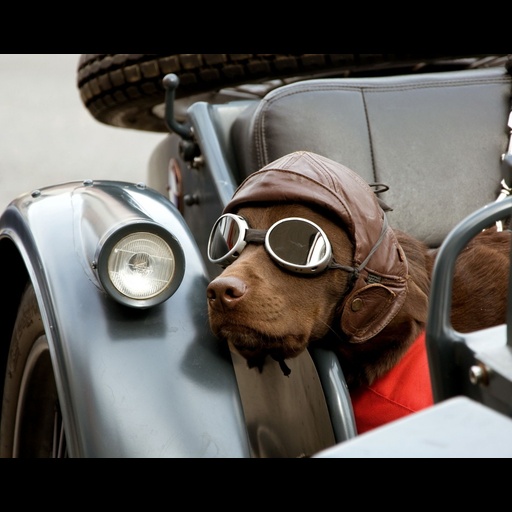} &
        \includegraphics[width=0.22\linewidth]{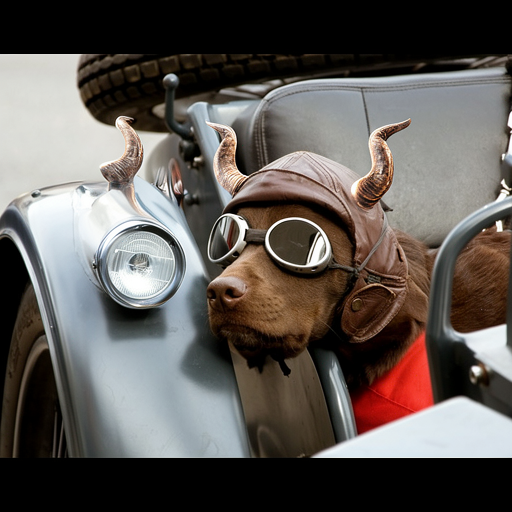} &
        \includegraphics[width=0.22\linewidth]{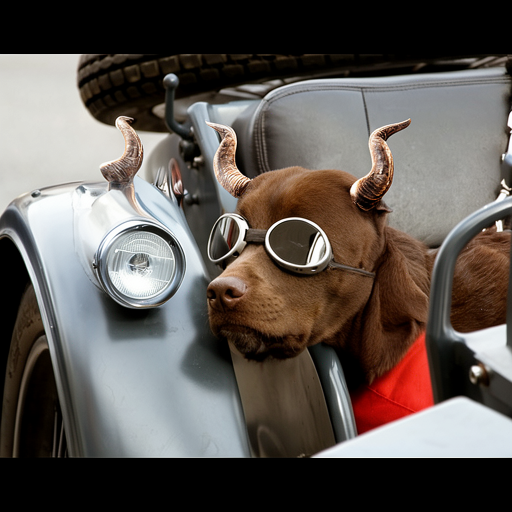} &
        \includegraphics[width=0.22\linewidth]{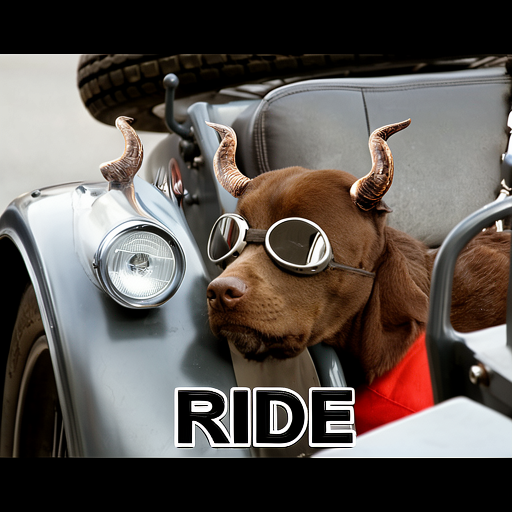} \\
        Input & Turn 1 & Turn 2 & Turn 3
    \end{tabular}

    \vspace{0.35em}
    \parbox{0.95\linewidth}{\footnotesize
    \textbf{Turn 1:} Add a horn above the metal silver headlight.\\
    \textbf{Turn 2:} Remove the leather brown hat.\\
    \textbf{Turn 3:} Add the text ``RIDE'' to the image.}

    \vspace{0.7em}
    \begin{tabular}{cccc}
        \includegraphics[width=0.22\linewidth]{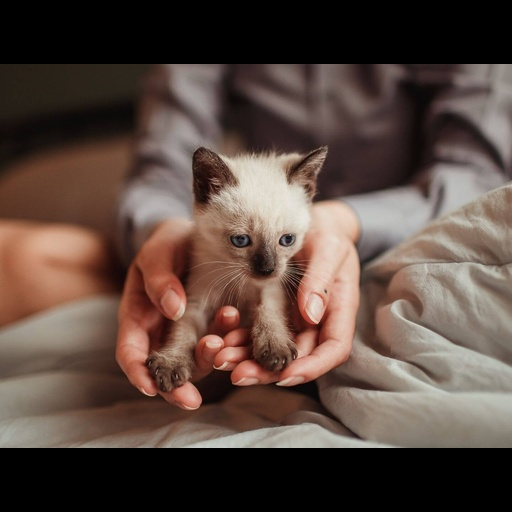} &
        \includegraphics[width=0.22\linewidth]{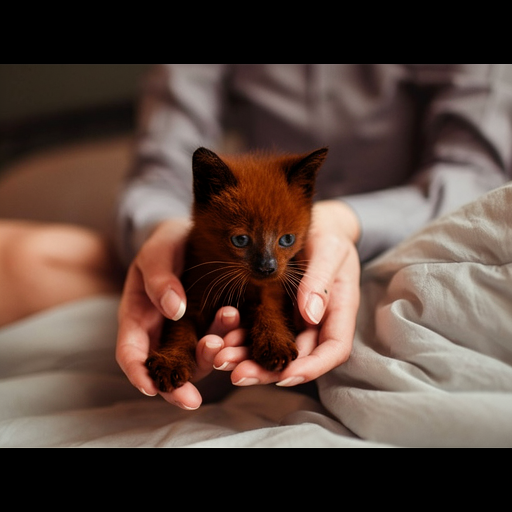} &
        \includegraphics[width=0.22\linewidth]{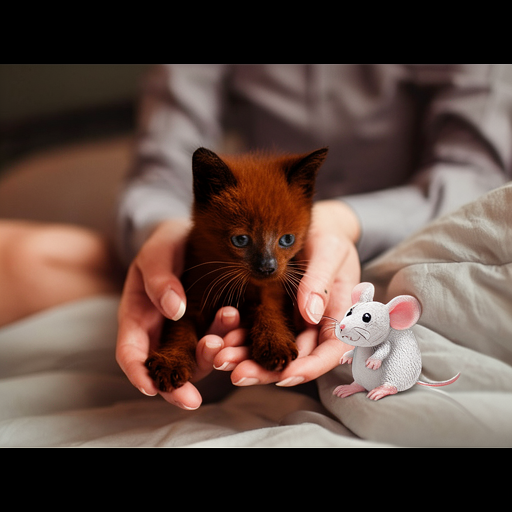} &
        \includegraphics[width=0.22\linewidth]{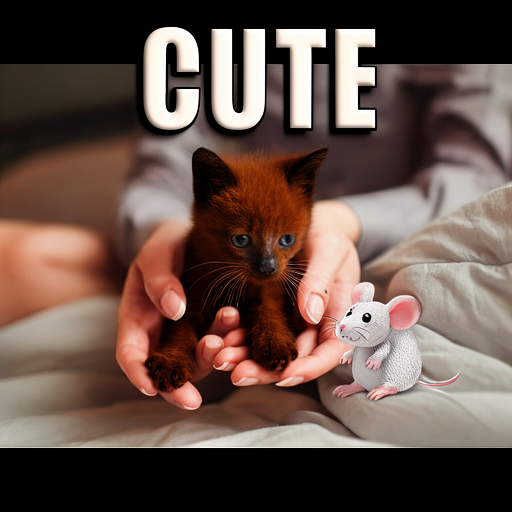} \\
        Input & Turn 1 & Turn 2 & Turn 3
    \end{tabular}

    \vspace{0.35em}
    \parbox{0.95\linewidth}{\footnotesize
    \textbf{Turn 1:} Change the color of fur cream kitten to brown.\\
    \textbf{Turn 2:} Add toy mouse on the right of brown fur cream kitten.\\
    \textbf{Turn 3:} Add text ``CUTE'' on the image.}
\end{figure*}

\end{document}